\documentclass{article} 
\usepackage{main,times}

\usepackage{amsmath,amsfonts,bm}

\def\eqref#1{equation~\ref{#1}}

\def\1{\bm{1}}

\DeclareMathAlphabet{\mathsfit}{\encodingdefault}{\sfdefault}{m}{sl}
\SetMathAlphabet{\mathsfit}{bold}{\encodingdefault}{\sfdefault}{bx}{n}

\newcommand{\OurMethod}{\emph{IC-Custom}}
\newcommand{\OurData}{\emph{CustomData}}
\newcommand{\OurBench}{\emph{ProductBench}}

\usepackage{booktabs} 
\usepackage[ruled]{algorithm2e} 
\usepackage{multirow}
\usepackage{pifont}
\usepackage{booktabs}
\usepackage{graphicx}
\definecolor{LightBlue}{rgb}{0.9,0.94,1}
\usepackage{fp}
\usepackage{adjustbox}
\usepackage{caption}
\usepackage{enumitem}
\usepackage{colortbl}
\usepackage{wrapfig}
\usepackage{subcaption}    
\usepackage{makecell, xcolor}
\usepackage{etoc}

\def\eg{\emph{e.g.}}

\usepackage{tikz}
\newcommand{\blackcircle}[1]{%
\tikz[baseline=(char.base),baseline=-0.7ex]{\node[shape=circle,fill=black,text=white,inner sep=0.5pt,font=\scriptsize] (char) {#1};}%
}

\usepackage{hyperref}
\usepackage{url}

\title{IC-Custom: Diverse Image Customization via In-Context Learning}

\author{\textbf{Yaowei Li}$^{1}$, \textbf{Xiaoyu Li}$^{2}$\thanks{Corresponding author}, 
\textbf{Zhaoyang Zhang}$^{2}$, \textbf{Yuxuan Bian}$^{4}$, 
\textbf{Gan Liu}$^{3}$, \textbf{Xinyuan Li}$^{3}$, \textbf{Jiale Xu}$^{2}$,
  \\
 \textbf{Wenbo Hu}$^{2}$, \textbf{Yating Liu}$^{5}$, \textbf{Lingen Li}$^{4}$, 
\textbf{Jing Cai}$^{3}$, \textbf{Yuexian Zou}$^{1}$\footnotemark[1], 
\textbf{Yancheng He}$^{3}$, \textbf{Ying Shan}$^{2}$ \\
\\
$^{1}$Peking University \quad
$^{2}$ARC Lab, Tencent PCG \quad
$^{3}$Tencent \\
$^{4}$The Chinese University of Hong Kong \quad
$^{5}$Tsinghua University \\
\\
\texttt{Project Page: \url{https://liyaowei-stu.github.io/project/IC_Custom/}}
}

\definecolor{citecolor}{RGB}{0,150,136}   
\definecolor{linkcolor}{RGB}{233,102,0}   
\definecolor{urlcolor}{RGB}{233,102,0}    
\hypersetup{
    colorlinks=true,
    citecolor=citecolor,
    linkcolor=linkcolor,
    urlcolor=urlcolor
}

\iclrfinalcopy 
\begin{document}

\maketitle

\begin{figure}[h!]
    \centering
    \vspace{-16pt}
    \includegraphics[width=0.88\linewidth]{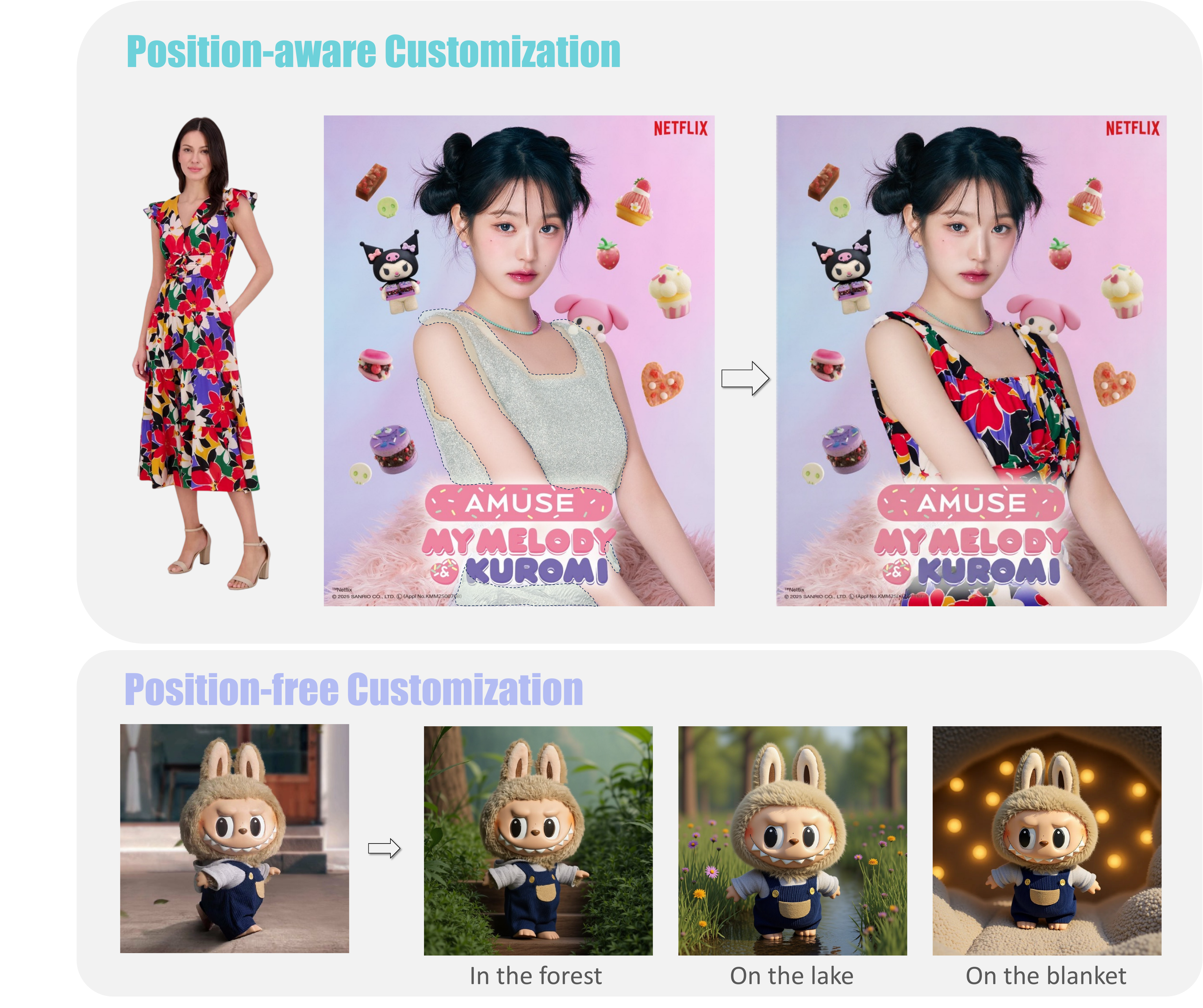}
    \vspace{-6pt}
    \captionof{figure}{\textbf{Visualization of \OurMethod{} results.} Our method supports diverse image customization scenarios, including position-aware (location-specified editing conditioned on a mask) and position-free (ID-consistent generation guided by text) customization.}
    \label{fig:teaser}
    \vspace{-2pt}
\end{figure}
\begin{abstract}
    Image customization, a crucial technique for industrial media production, aims to generate content that is consistent with reference images.
    However, current approaches conventionally separate image customization into position-aware and position-free customization paradigms and lack a universal framework for diverse customization, limiting their applications across various scenarios.
    To overcome these limitations, we propose \OurMethod{}, a unified framework that seamlessly integrates position-aware and position-free image customization through in-context learning. \OurMethod{} concatenates reference images with target images to a polyptych, leveraging DiT's multi-modal attention mechanism for fine-grained token-level interactions. 
    We propose the In-context Multi-Modal Attention (ICMA) mechanism, which employs learnable task-oriented register tokens and boundary-aware positional embeddings to enable the model to effectively handle diverse tasks and distinguish between inputs in polyptych configurations.
    To address the data gap, we curated a 12K identity-consistent dataset with 8K real-world and 4K high-quality synthetic samples, avoiding the overly glossy, oversaturated look typical of synthetic data.
    \OurMethod{} supports various industrial applications, including try-on, image insertion, and creative IP customization. 
   Extensive evaluations on our proposed \OurBench{} and the publicly available \emph{DreamBench} demonstrate that \OurMethod{} significantly outperforms community workflows, closed-source models, and state-of-the-art open-source approaches. \OurMethod{} achieves about 73\% higher human preference across identity consistency, harmony, and text alignment metrics, while training only 0.4\% of the original model parameters.
 \end{abstract}

\section{Introduction}

Image customization, which ensures that generated content remains consistent with the identity of reference images, has enabled applications such as image insertion~\citep{chen2024zero, chen2024anydoor, mao2025ace++, song2025insert}, IP creation~\citep{ruiz2023dreambooth, ye2023ip, tewel2024training, ominicontrol, mou2025dreamo}, and visual try-on~\citep{wang2024stablegarment, guo2025any2anytryon, xu2025ootdiffusion}. These capabilities are vital for industrial media production, supporting consistent content creation across diverse visual contexts.

Early image customization methods~\citep{dreambooth, gal2022image, avrahami2023break} relied on per-instance optimization, which was time-consuming.  
Subsequent approaches~\citep{ipadapter, chen2024anydoor, chen2024zero} added control branches to pre-trained diffusion models to inject identity information from reference images. However, these methods were constrained by model architecture and scalability issues, resulting in suboptimal performance.
Recently, by leveraging the long-range modeling inductive bias of DiT architectures~\citep{peebles2023scalable, sd3, Flux}, image conditions can be directly input as sequences, interacting with noisy tokens through multi-modal attention mechanisms, without the need for additional branches. This enables image customization methods to exhibit powerful emergent capabilities~\citep{song2025insert, mou2025dreamo, FluxFill, FluxRedux, ominicontrol, mao2025ace++}.

\definecolor{darkgreen}{rgb}{0.0, 0.8, 0.0}
\definecolor{darkred}{rgb}{0.8, 0.0, 0.0}

\begin{wraptable}{r}{0.41\linewidth}
    \vspace{-12pt}
    \scriptsize
    \caption{Comparison of \OurMethod{} with previous image customization methods~\citep{FluxFill,FluxRedux,ominicontrol,song2025insert,mou2025dreamo, hurst2024gpt}. The checkmarks and crosses indicate task compatibility.}
    \vspace{-6pt}
    \renewcommand\arraystretch{.88}
    \setlength{\tabcolsep}{1mm}{
    \begin{tabular}{lccc}
    \toprule
    \multirow{2}{*}{\textbf{Model}} & \multicolumn{2}{c}{\textbf{Position-aware}} & \multirow{2}{*}{\textbf{Position-free}} \\
    \cmidrule(lr){2-3}
    & \emph{precise}& \emph{user-drawn}& \\
    \midrule
    FLUX.1 workflow & \textcolor{darkgreen}{\ding{51}} & \textcolor{darkgreen}{\ding{51}} & \textcolor{darkred}{\ding{55}} \\
    OminiCtrl & \textcolor{darkred}{\ding{55}} & \textcolor{darkred}{\ding{55}} & \textcolor{darkgreen}{\ding{51}} \\
    Insert Anything & \textcolor{darkgreen}{\ding{51}} & \textcolor{darkgreen}{\ding{51}} & \textcolor{darkred}{\ding{55}} \\
    DreamO & \textcolor{darkred}{\ding{55}} & \textcolor{darkred}{\ding{55}} & \textcolor{darkgreen}{\ding{51}} \\
    GPT-4o & \textcolor{darkred}{\ding{55}} & \textcolor{darkred}{\ding{55}} & \textcolor{darkgreen}{\ding{51}} \\
    \midrule
    \OurMethod & \textcolor{darkgreen}{\ding{51}} & \textcolor{darkgreen}{\ding{51}} & \textcolor{darkgreen}{\ding{51}} \\
    \bottomrule
    \end{tabular}
    }
    \label{tab:compare_other_methods}
    \vspace{-6pt}
\end{wraptable}

Despite these advances, existing methods still face significant challenges in maintaining consistent identity across diverse user requirements and customization scenarios (see Tab.~\ref{tab:compare_other_methods}):
\textbf{(1)} They typically treat image customization as two separate tasks. In \emph{position-aware} customization, an reference identity is inserted into masked regions of a fill-in image. In \emph{position-free} customization, identity-consistent images are generated from text prompts.
\textbf{(2)} They provide limited support for diverse mask types, often confusing user-drawn with precise masks, \eg, treating coarse hand-drawn regions as exact boundaries.
These limitations hinder the development of unified frameworks capable of flexibly handling diverse customization requirements, forcing separate models for each scenario and limiting the development of robust, comprehensive identity representations.

To this end, we propose \OurMethod{}, a unified framework that seamlessly integrates position-aware and position-free image customization, enabling flexible and identity-consistent customization across diverse scenarios (see Fig.~\ref{fig:teaser}). 
Specifically, we first employ a diptych format by concatenating the reference identity image with the fill-in image (either partially or fully masked), yielding a unified representation that allows the model to handle diverse customization settings within a single framework. Building on DiT's multi-modal attention, we further introduce a novel In-Context Multi-Modal Attention (ICMA) module that more effectively transfers identity information from the reference image to the fill-in image and enables comprehensive customization across diverse scenarios.
The ICMA module features two key innovations: 
\textbf{(1)} Three types of learnable, task-oriented register tokens to specify the customization type—position-aware customization (with precise or user-drawn masks) and position-free customization—allowing the model to adapt its behavior based on user requirements.
\textbf{(2)} Two types of learnable positional embeddings to represent spatial relationships: Reference Embeddings (RE) for the reference identity image and Fill Embeddings (FE) for the fill-in image, helping the model clearly differentiate input boundaries in the diptych format.

To enable effective training of our unified framework, we curated a high-quality dataset \OurData{}, consisting of both real-world and synthetic samples. Specifically, we curated 8K identity-consistent diptychs from real-world sources and an additional 4K synthetic diptychs, resulting in a total of 12K diptychs. This comprehensive dataset enables our model to learn robust identity representations across diverse contexts and viewpoints, while also addressing the limitations of previous methods that overly rely on synthetic data and often produce artificial-looking results.

To extensively evaluate the performance of our method, we use \OurBench{} and \emph{DreamBench}~\citep{dreambooth} to assess both position-aware and position-free customization capabilities. \OurBench{} is our manually curated benchmark for position-aware customization, consisting of 40 identity-consistent images with an even distribution of rigid and non-rigid objects, along with their corresponding precise and user-drawn masks. We also use DreamBench to evaluate position-free customization performance. Extensive subjective and objective evaluations demonstrate that \OurMethod{} outperforms community workflows, the closed-source GPT-4o (March 25, 2025), and state-of-the-art open-source approaches. Notably, \OurMethod{} achieves a 73\% higher human preference across identity consistency, harmony, and text alignment metrics, while training only 0.3\% of the parameters of the pre-trained FLUX model.

In summary, our contributions are as follows:
\begin{itemize}[leftmargin=*,itemsep=-0.1em]
    \item We propose a unified framework that seamlessly integrates position-aware and position-free image customization via in-context formulation.
    \item We introduce the ICMA module, which enables flexible image customization through learnable task-oriented register tokens and boundary-aware positional embeddings.
    \item We curate a dataset from real-world sources, addressing the limitations of existing methods that rely on synthetic data, which often produce  artificial-looking results.
    \item We demonstrate that our method outperforms existing approaches across a range of metrics, surpassing community workflows, closed-source models, and state-of-the-art open-source methods.
\end{itemize}

\section{Preliminaries}

\paragraph{\textbf{MM-DiT Architecture.}}
Recent state-of-the-art generative diffusion models, such as SD3~\citep{sd3} and FLUX~\citep{Flux}, leverage the MM-DiT architecture~\citep{dit}, which integrates a Multi-modal Attention (MMA) mechanism with Rotary Position Embedding (RoPE) as a central component. This design enables the concurrent processing of noisy image tokens $X_t \in \mathbb{R}^{n \times d}$ and text tokens $C_{\mathrm{T}} \in \mathbb{R}^{l \times d}$, as shown in Eq.~\ref{eq:mm-dit-mma}.

\begin{equation}
\label{eq:mm-dit-mma}
\operatorname{MMA}\left(\left[X_t ; C_{\mathrm{T}}\right]\right) = \operatorname{softmax}\left(\frac{\mathcal{R}(Q) \cdot \mathcal{R}(K)^{\top}}{\sqrt{d}}\right) \mathcal{R}(V).
\end{equation}
Here, $Q$, $K$, and $V$ are derived from the projection of the concatenated input $[X_t; C_{\mathrm{T}}] \in \mathbb{R}^{(n+l) \times d}$, with the operator $\mathcal{R}$ applying RoPE to $Q$ and $K$ to encode positional information.

\paragraph{Flow Matching.}
The model is trained within the Rectified Flow (RF)~\citep{rectified_flow}. The Continuous Normalizing Flow (CNF) is formalized as the following ODE:
\begin{equation}
    \frac{d}{d t} X_t = v(X_t,t)dt = X_1 - X_0, \quad \forall t \in [0, 1].
\end{equation}
Here, given a clean latent variable $X_0 \sim p_{\text{data}}$ and a Gaussian noise sample $X_1 \sim \mathcal{N}(0, 1)$, $X_t$ is constructed via linear interpolation:
\begin{equation}
    \label{eq:linear_interpolation}
    X_t = t X_1 + (1 - t) X_0, \quad \forall t \in [0, 1].
\end{equation}

Subsequently, the Conditional Flow Matching (CFM) loss~\citep{lipman2023flow} is employed to train a velocity filed prediction model  $v_{\Theta}$:
\begin{equation}
    \mathcal{L}_{\mathrm{CFM}} 
    =
    \mathbb{E}_{\,t \sim p(t),\, X_{1} \sim \mathcal{N}(0,1),\, (X_{0},C_{\mathrm{T}}) \sim p_{\text{data}}}
    \Bigl[
    \bigl\|
    v_{\Theta}\!\left(X_{t}, C_{\mathrm{T}}, t\right)
    - \bigl(X_{1} - X_{0}\bigr)
    \bigr\|_{2}^{2}
    \Bigr].
\end{equation}

Here, $t$ is sampled from a \emph{Logit-Normal Distribution}~\citep{esser2024scaling} with the probability density function $p(t) = \frac{\exp(-0.5 \cdot (\mathrm{logit}(t) - \mu)^2 / \sigma^2)}{\sigma \sqrt{2\pi} \cdot (1 - t) \cdot t}$, where $\mathrm{logit}(t) = \log\frac{t}{1 - t}$. From the Logit-Normal Distribution definition, $Y = \mathrm{logit}(t) \sim \mathcal{N}(\mu, \sigma)$, with $\mu = 0$ and $\sigma = 1$ under the RF.

\paragraph{DiT-based Image Customization Methods}

Recent state-of-the-art DiT-based image customization methods~\citep{unireal,mao2025ace++,wu2025less,song2025insert,mou2025dreamo}, integrate reference image conditions directly into the input via concatenation, instead of using additional network branches. This method unifies reference and other conditions into a single sequence, improving integration during flow matching. 
However, these methods typically train position-aware and position-free customization tasks separately, without explicitly addressing their potential unification. In position-aware tasks, the identity's location is specified using a mask, while position-free tasks leverage textual guidance to generate identity-consistent content. For instance, ACE++~\citep{mao2025ace++} and OmniControl~\citep{ominicontrol} train separate LoRA adapters, InsertAnything~\citep{song2025insert} is specifically trained for position-aware tasks, and DreamO~\citep{mou2025dreamo} and UNO~\citep{wu2025less} are designed for position-free tasks.
\section{Method}
\begin{figure*}[t]
    \centering
    \includegraphics[width=\linewidth]{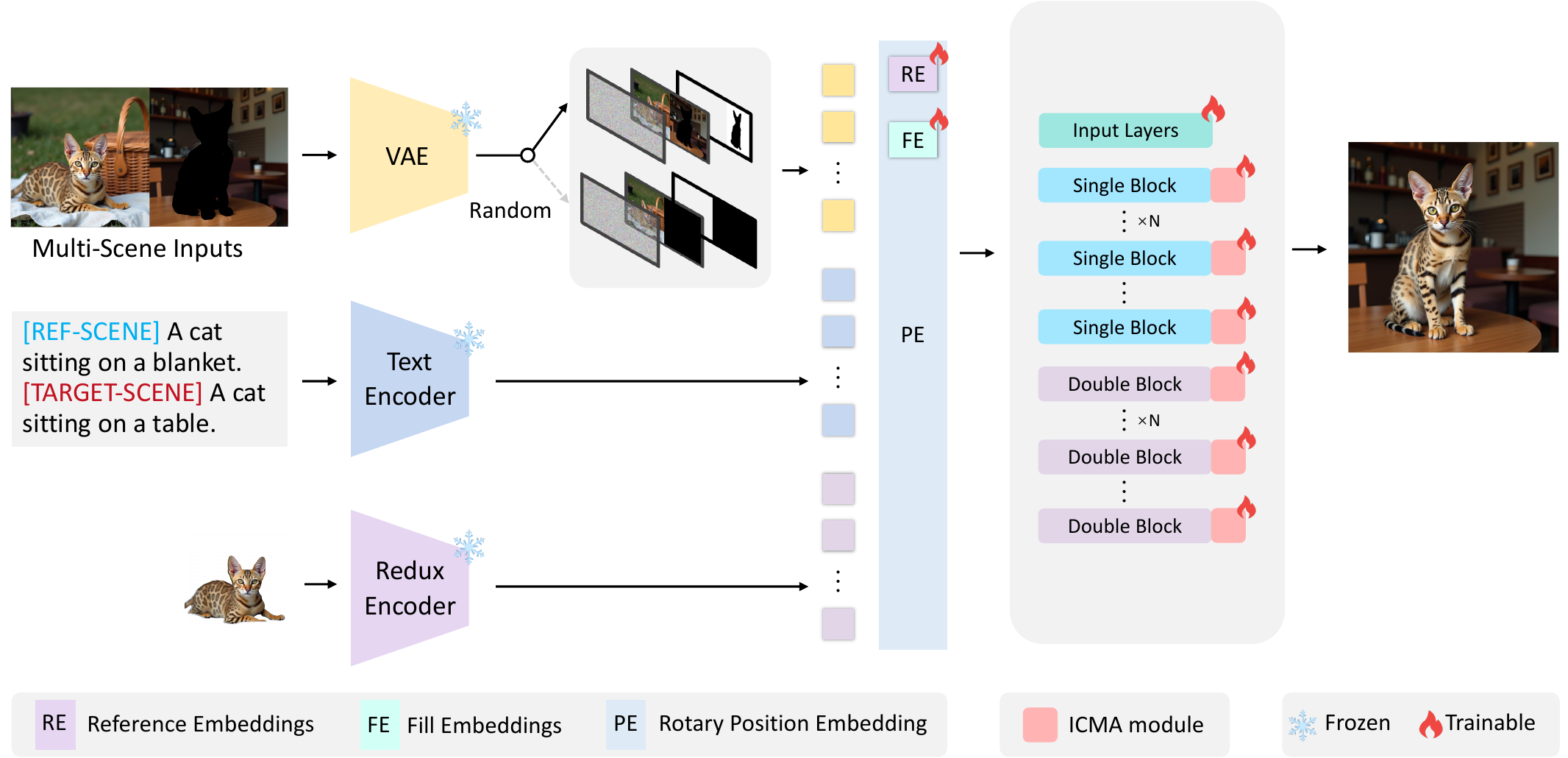}
    \caption{\textbf{Model overview.} 
        (1) Our model takes in-context diptych inputs together with redux embeddings and text prompts. 
        (2) During training, it randomly chooses to mask either the entire fill-in image (position-free customization) or only partial regions (position-aware customization) to produce diverse in-context latents. 
        (3) The ICMA module, equipped with task-oriented register tokens and boundary-aware positional embeddings (see Sec.~\ref{sec:in-context-multi-modal-attention}), is integrated into the architecture. 
        We train LoRA adapters on the ICMA module while unfreezing the input layers.
        }
    \label{fig:model}
\end{figure*}

As shown in Fig.~\ref{fig:model}, we introduce \OurMethod{}, a novel approach that presents a unified framework for comprehensive image customization, as detailed in Sec.~\ref{sec:in-context-diptych-customization}. At its core, \OurMethod{} leverages In-Context Multi-Modal Attention (ICMA) to effectively adapt to diverse customization scenarios, as described in Sec.~\ref{sec:in-context-multi-modal-attention}. Additionally, we curate a high-quality dataset for comprehensive customization tasks, sourced from both real-world and synthetic data, with image resolutions exceeding 800×800 pixels, as outlined in Sec.~\ref{sec:in-context-customization-data-curation}.

\subsection{In-Context Diptych Customization}
\label{sec:in-context-diptych-customization}
\paragraph{\textbf{Motivation.}} 
Formally, position-aware customization can be framed as a reference-guided image filling task, represented as $p(\hat{X} \mid C_{\mathrm{I}}, C_{\mathrm{I}'}, M)$, where $\hat{X}$ denotes the customized output, $C_{\mathrm{I}}$ denotes the reference identity image, $C_{\mathrm{I}'}$ represents the image to be filled, and $M$ denotes the mask specifying the filling position.
In contrast, position-free customization is viewed as a reference-guided text-to-image task, formalized as $p(\hat{X} \mid C_{\mathrm{I}}, C_{\mathrm{T}})$. 
Since position-free customization can be regarded as a special case of image filling where $M$ and $C_{\mathrm{I}'}$ are set to zero, we unify both paradigms under the formulation $p(\hat{X} \mid C_{\mathrm{I}}, C_{\mathrm{I}'}, M, C_{\mathrm{T}})$. 

\paragraph{\textbf{Diptych Framework and Training Strategy.}} 
Based on the unified formulation above, we introduce an in-context diptych format to unify diverse input conditions and support this paradigm.
Specifically, we concatenate the reference identity image $C_{\mathrm{I}}$ with the fill-in image $C_{\mathrm{I}'}$ in a diptych layout, then encode them jointly as tokens to enforce simultaneous modeling and generation. 
The model is trained with the following CFM loss:
\begin{equation}
   \label{eq:velocity-field-prediction-objective-diptych}
    \mathcal{L}_{\mathrm{CFM}} 
    =
    \mathbb{E}_{\,t \sim p(t),\, X_{1} \sim \mathcal{N}(0,1),\, (X_{0}, C_{\mathrm{T}}) \sim p_{\text{data}}}
    \Bigl[
    \bigl\|
    v_{\Theta}\!\left([X_{t},X_0^{m},M], C_{\mathrm{T}}, t\right)
    - \bigl(X_{1} - X_{0}\bigr)
    \bigr\|_{2}^{2}
    \Bigr],
\end{equation}
where $X_{0} = [C_{\mathrm{I}}; C_{\mathrm{I}'}]$ denotes the width-wise diptych concatenation of the reference identity image and the fill-in image, $X_{t}$ is computed according to Eq.~\ref{eq:linear_interpolation}, and $X_{0}^{m} = X_{0} \odot M$, with $\odot$ indicating element-wise multiplication. 
Here, $[X_{t}, X_{0}^{m}, M]$ represents the channel-wise concatenation of these three components.
The text condition $C_{\mathrm{T}}$ provides scene descriptions for both the reference identity image and the fill-in image, separated by the placeholders [REF-SCENE] and [TARGET-SCENE]. 
Notably, during training, instead of requiring triplets $(C_{\mathrm{I}}, C_{\mathrm{I}'}, \hat{X})$, where $C_{\mathrm{I}'}$ and $\hat{X}$ typically differ in identity, we use two images of the same identity and set $\hat{X}=C_{\mathrm{I}'}$, enabling the model to predict $C_{\mathrm{I}'}$ conditioned on $M$ and $X_{0}^{m}$; hence Eq.~\ref{eq:velocity-field-prediction-objective-diptych} defines $X_{0}=[C_{\mathrm{I}};C_{\mathrm{I}'}]$ rather than $X_{0}=[C_{\mathrm{I}};\hat{X}]$.

Based on this formulation, once paired data $\{C_{\mathrm{I}},C_{\mathrm{I}'},M,C_{\mathrm{T}}\}$ are available, the model can be trained in two complementary modes without collecting separate datasets or designing distinct model structures. 
Specifically, setting $C_{\mathrm{I}'}$ and $M$ to zero (i.e., a global mask) corresponds to position-free customization, while using nonzero (localized) masks for $C_{\mathrm{I}'}$ and $M$ enables position-aware customization. 
Thus, a single paired dataset suffices to support both capabilities through simple variations in training inputs.

In implementation, as shown in Fig.~\ref{fig:model}, we use a VAE~\citep{vae} to encode the input diptych, while T5~\citep{raffel2020exploring} and CLIP~\citep{clip} serve as text encoders for the text prompts. Optionally, FLUX.1 Redux~\citep{FluxRedux} is employed to further encode identity information. The resulting representations are then fed into DiT blocks equipped with the ICMA module (see Sec.~\ref{sec:in-context-multi-modal-attention}) for flow matching.

\subsection{In-Context Multi-Modal Attention}
\label{sec:in-context-multi-modal-attention}
\paragraph{\textbf{Challenges.}} 
Although our pipeline seamlessly adapts to diverse customization settings, it still faces several challenges. 
\textbf{(1)} \emph{Task-type ambiguity}: for example, under position-aware customization settings, the model often misinterprets user-drawn masks as precise boundaries, generating content that fully fills and strictly follows the mask shape. 
\textbf{(2)} \emph{Image-boundary confusion}: in diptych prediction settings (Eq.~\ref{eq:velocity-field-prediction-objective-diptych}), the model struggles to differentiate between reference and target regions, leading to undesirable edge artifacts.

\begin{figure*}[t]
  \centering
  \includegraphics[width=0.9\textwidth]{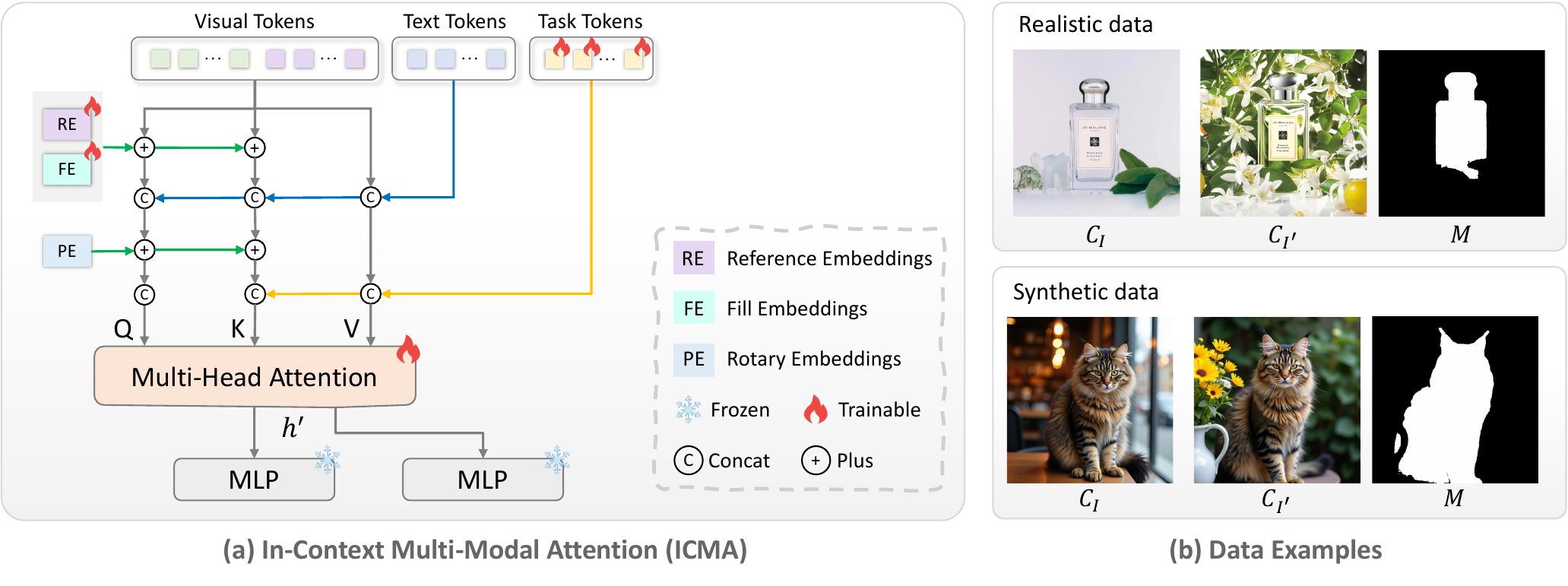}
 \caption{\textbf{(a) In-Context Multi-Modal Attention (ICMA).} 
ICMA incorporates learnable task-oriented register tokens and boundary-aware positional embeddings (RE, FE) into the multi-modal attention of MM-DiT~\citep{dit} to specify customization types and delineate input boundaries. 
\textbf{(b) Training data examples.} High-quality identity-consistent quadruples $\{C_{\mathrm{I}}, C_{\mathrm{I}'}, M, C_{\mathrm{T}}\}$ from real-world and synthetic data; for clarity, text descriptions $C_{\mathrm{T}}$ are omitted.
  }
  \label{fig:icma}
\end{figure*}

\paragraph{\textbf{Proposed ICMA.}} 
To address these issues, we propose In-Context Multi-Modal Attention module (ICMA), a variant of the multi-modal attention mechanism. 
As illustrated in Fig.~\ref{fig:icma} (a), ICMA incorporates two key design innovations: 
\textbf{(1)} \emph{learnable task-oriented register tokens} to explicitly indicate the customization type (precise masks, user-drawn masks, or position-free); and 
\textbf{(2)} \emph{learnable boundary-aware positional embeddings}—comprising Reference Embeddings (RE) and Fill Embeddings (FE)—to encode spatial relationships between the reference identity image and the fill-in image.
Formally, the ICMA mechanism operates as follows:
\begin{equation}
\label{eq:icma}
\begin{aligned}
   \mathcal{P}(x) &= x + [\,\mathcal{E}_{\mathrm{R}};\,\mathcal{E}_{\mathrm{F}}\,] + \mathcal{R}(x), \\
   Q &= [\,\mathcal{P}(Q_{\mathrm{I}});\; Q_{\mathrm{T}} + \mathcal{R}(Q_{\mathrm{T}})\,], \\
   K &= [\,\mathcal{P}(K_{\mathrm{I}});\; K_{\mathrm{T}} + \mathcal{R}(K_{\mathrm{T}})\,];\; \mathbf{r}_{i}\,], \\
   V &= [\,V_{\mathrm{I}};\; V_{\mathrm{T}};\; \mathbf{r}_{i}\,], \\
   h' &= \mathrm{MHA}(Q,K,V),
\end{aligned}
\end{equation}
where $[;]$ denotes diptych concatenation, $\mathcal{R}(\cdot)$ denotes rotary position encoding~\citep{su2024roformer}; 
$Q_{I},K_{I},V_{I}\in\mathbb{R}^{n\times d}$ and $Q_{T},K_{T},V_{T}\in\mathbb{R}^{l\times d}$ are the query, key, and value matrices for image and text tokens, respectively; 
$\mathcal{E}_{\mathrm{R}},\mathcal{E}_{\mathrm{F}}$ are the learnable Reference and Fill embeddings; 
$\mathbf{r}_{i}\in\mathbb{R}^{m\times d}$ denotes the $i$-th learnable task-oriented register token; 
and $\mathrm{MHA}(\cdot)$ is the Multi-Head Attention operation.
Our proposed ICMA module replaces the multi-modal attention layers in both the double-block and single-block components of the original FLUX.1 MM-DiT architecture~\citep{Flux}.

\subsection{In-Context Customization Data Curation}
\label{sec:in-context-customization-data-curation}
\paragraph{\textbf{Data Collection.}}  
The scarcity of high-quality customization data remains a critical bottleneck in developing robust customization models. 
Existing approaches~\citep{ominicontrol, wu2025less, li2025visualcloze} rely predominantly on synthetic data for training; however, such data often struggles to preserve identity consistency and photorealistic quality, thereby limiting model effectiveness.

To address this challenge, we introduce \OurData{}, a high-quality customization dataset designed for both authenticity and diversity. 
We curate nearly 8K identity-consistent realistic image pairs from e-commerce platforms, covering real-world scenarios such as clothing try-on, cosmetics, furniture, electronics, accessories, home decor, and personal care products, with resolutions ranging from $800\times800$ to $3000\times3664$ pixels. 
To further enrich the dataset and extend coverage beyond commercial products, we add 4K high-quality, identity-consistent synthetic pairs carefully filtered from the SynCD $1024 \times 1024$ subset~\citep{kumari2025generating}, resulting in a comprehensive dataset of 12K $\{C_{\mathrm{I}}, C_{\mathrm{I}'}, M, C_{\mathrm{T}}\}$ samples (see Fig.~\ref{fig:icma}(b) for visualization; symbol definitions in Sec.~\ref{sec:in-context-diptych-customization}).

\paragraph{\textbf{Data Processing.}}  
Our filtering process applies three rules:  
(1) exclude items whose DINOv2~\citep{oquab2023dinov2} feature similarity between $C_{\mathrm{I}}$ and $C_{\mathrm{I}'}$ is below 0.2;  
(2) discard pairs composed entirely of blank-background images; and  
(3) ensure $C_{\mathrm{I}'}$ is not a blank-background image.  
These rules improve identity consistency and reduce ambiguity.  
We then use Qwen-VL2.5~\citep{bai2025qwen2} to auto-generate captions for \OurData{} (system prompt in Appendix Sec.~\ref{sec:automated_caption}) and Grounded SAM~\citep{ren2024grounded} to obtain ground-truth masks, while randomly generating user masks under predefined rules to support model training (see Appendix Sec.~\ref{appendix:mask_sampling_augmentation} for details).

\section{Experiments}
\subsection{Experiments Setup}
\label{sec:expriments_setup}
\paragraph{\textbf{Implementation Details.}}
\OurMethod{} builds on the pre-trained text-to-image model FLUX.1-Fill~\citep{FluxFill}. We train LoRA~\citep{hu2022lora} (rank 64) on the first 10 layers of both single and double blocks, while directly fine-tuning the image and text input layers. In total, only $49.26$M parameters are trainable—just $0.4\%$ of the original FLUX model’s $12$B parameters (19 double and 38 single blocks). 
Unlike prior methods~\citep{song2025insert,mou2025dreamo} that train LoRA on all layers (e.g., DreamO~\citep{mou2025dreamo} trained 707M parameters), our approach drastically cuts training cost. The model is optimized on our 12K dataset for 20K iterations using AdamW~\citep{adam} with a learning rate of $5\times10^{-5}$ and a batch size of 4. To handle diverse resolutions, we employ a data-bucketing strategy that groups samples by size (e.g., $800{\times}800$, $1024{\times}1024$, $1024{\times}1280$, $1280{\times}1280$, $1504{\times}1504$) so each batch has uniform input dimensions. We also present a web application and inference pipeline in Appendix~\ref{appendix:web_app}.

\paragraph{\textbf{Benchmarks.}}
To assess our model's performance in both position-aware and position-free customization settings, we evaluate on our proposed \OurBench{} and the open-source \emph{DreamBench}~\citep{dreambooth} benchmark. 
\OurBench{} contains 40 high-quality, identity-consistent items with resolutions exceeding $1024\times1024$ pixels. Each item includes paired images and corresponding masks, with no overlap with our training data. We use SAM~\citep{kirillov2023segment} to annotate precise masks and manually create user-drawn masks. The dataset is evenly divided into rigid and non-rigid categories, covering diverse domains such as clothing try-on, accessories, bags, furniture, toys, and perfume, specifically designed to evaluate position-aware customization. 
\emph{DreamBench} comprises 30 items, each with 5–6 identity-consistent images and used to evaluate position-free customization. We take the first image of each item as the reference.
Additionally, we use Qwen-VL2.5~\citep{bai2025qwen2} to generate in-context textual descriptions for both benchmarks. For \OurBench{}, we directly prompt it to caption the diptych input, whereas for \emph{DreamBench} we prompt it to creatively generate new scene descriptions. (see Appendix Sec.~\ref{sec:automated_caption_for_benchmark} for details)

\begin{table}[t]
  \small
  \setlength\tabcolsep{5pt}
  \renewcommand\arraystretch{1.02}
  \caption{\textbf{Quantitative results on position-aware and position-free image customization.} 
  Evaluation on \OurBench{} (precise/user-drawn masks) and \emph{DreamBench} shows that \OurMethod{} consistently outperforms existing methods across all objective metrics (higher is better $\uparrow$). 
  Baselines: FLUX.1 workflow~\citep{FluxFill,FluxRedux}, OminiCtrl/DreamO/Insert Anything~\citep{ominicontrol,mou2025dreamo,song2025insert}, GPT-4o~\citep{hurst2024gpt}.}
  \label{tab:objective_comparisons}
  \resizebox{\columnwidth}{!}{
  \begin{tabular}{l|ccc|ccc|ccc}
    \toprule
    \multirow{3}{*}{\textbf{Method}} &
    \multicolumn{6}{c|}{\textbf{ProductBench}} &
    \multicolumn{3}{c}{\textbf{DreamBench}} \\
    \cmidrule(lr){2-7}\cmidrule(lr){8-10}
    & \multicolumn{3}{c|}{\emph{Precise Mask}} 
    & \multicolumn{3}{c|}{\emph{User-drawn Mask}} 
    & \multicolumn{3}{c}{\emph{Position-free}} \\
    \cmidrule(lr){2-4}\cmidrule(lr){5-7}\cmidrule(lr){8-10}
    & DINO-I $\uparrow$ & CLIP-I $\uparrow$ & CLIP-T $\uparrow$
    & DINO-I $\uparrow$ & CLIP-I $\uparrow$ & CLIP-T $\uparrow$
    & DINO $\uparrow$ & CLIP $\uparrow$ & CLIP-T $\uparrow$ \\
    \midrule
    FLUX.1 workflow & 60.80 & \underline{81.66} & 31.13 & \underline{62.26} & 81.60 & 31.29 & — & — & — \\
    OminiCtrl       & 57.93 & 76.06 & \underline{31.31} & — & — & — & 48.29 & 75.85 & \underline{36.82} \\
    DreamO          & \underline{62.98} & 78.86 & 31.25 & — & — & — & \underline{57.69} & 76.33 & 36.24 \\
    Insert Anything & 62.71 & 81.65 & 31.24 & 61.21 & \underline{81.75} & \underline{31.44} & — & — & — \\
    GPT-4o          & 61.40 & 78.53 & 30.72 & 62.05 & 79.87 & 30.58 & 54.31 & \underline{77.38} & 36.33 \\
    \rowcolor{blue!5}\textbf{IC-Custom (Ours)} & 
    \textbf{63.14} & \textbf{81.92} & \textbf{31.75} & 
    \textbf{63.28} & \textbf{81.95} & \textbf{31.80} & 
    \textbf{65.67} & \textbf{83.19} & \textbf{36.88} \\
    \bottomrule
  \end{tabular}}
  \vspace{-4pt}
\end{table}

\begin{table}[t]
  \centering
  \captionsetup{skip=5pt}
    \caption{
    (a) \textbf{Human-study results on image customization quality (higher is better)}. 
    (b) \textbf{Ablation studies on ProductBench.}
    Abbreviations: Zero-shot = zero-shot inference without fine-tuning; 
    w/o IL = without training Input Layers; 
    w/o RD = without using Real Data for training; 
    w/o UM = without using User-drawn Mask for training; 
    w/o TR = without Task-oriented Register tokens; 
    w/o PE = without Boundary-aware Positional Embeddings.}
  \label{tab:human-ablation-side}
  \scriptsize
  \setlength\tabcolsep{3pt}

  \begin{minipage}[t]{0.48\linewidth}
    \centering
    \textbf{(a) Human-study results}\\
    \renewcommand{\arraystretch}{1.28} 
    \resizebox{\linewidth}{!}{
    \begin{tabular}{l|ccc}
      \toprule
      \textbf{Method} & \textbf{Consistency $\uparrow$} & \textbf{Harmony $\uparrow$} & \textbf{Text Alignment $\uparrow$} \\
      \midrule
      FLUX.1 workflow & 3.2\% & 5.3\% & — \\
      OminiCtrl       & 1.5\% & 2.1\% & 6.3\% \\
      DreamO          & 5.4\% & 3.2\% & 10.1\% \\
      Insert Anything & 6.8\% & 6.5\% & — \\
      GPT-4o          & 4.6\% & 7.5\% & 21.4\% \\
      \rowcolor{blue!5}\textbf{IC-Custom (Ours)} & \textbf{78.5\%} & \textbf{75.4\%} & \textbf{62.2\%} \\
      \bottomrule
    \end{tabular}}
    \renewcommand{\arraystretch}{1.05}
  \end{minipage}\hfill
  \begin{minipage}[t]{0.5\linewidth}
    \centering
    \textbf{(b) Ablation on ProductBench}\\
    \resizebox{\linewidth}{!}{
    \begin{tabular}{l|ccc|ccc}
      \toprule
      \textbf{Models} & \multicolumn{3}{c|}{\textbf{Precise Mask}} & \multicolumn{3}{c}{\textbf{User-drawn Mask}} \\
      \cmidrule(lr){2-4}\cmidrule(lr){5-7}
      & \textbf{DINO-I $\uparrow$} & \textbf{CLIP-I $\uparrow$} & \textbf{CLIP-T $\uparrow$}
      & \textbf{DINO-I $\uparrow$} & \textbf{CLIP-I $\uparrow$} & \textbf{CLIP-T $\uparrow$} \\
      \midrule
      Zero-shot & 55.49 & 77.55 & 31.24 & 57.63 & 79.84 & 31.20 \\
      w/o IL    & 62.00 & 81.52 & 31.36 & 62.13 & 81.33 & 31.64 \\
      w/o RD    & 62.38 & 81.81 & 31.62 & 62.71 & 81.85 & 31.22 \\
      w/o UM    & 62.65 & 81.82 & 31.58 & 61.30 & 81.28 & 31.64 \\
      w/o TR    & 63.00 & 81.42 & 31.43 & 63.07 & 81.44 & 31.33 \\
      w/o PE    & 62.99 & 81.31 & 31.42 & 63.08 & 81.40 & 31.30 \\
      \midrule
      \rowcolor{blue!5}\textbf{Ours} & \textbf{63.14} & \textbf{81.92} & \textbf{31.75} & \textbf{63.28} & \textbf{81.95} & \textbf{31.80} \\
      \bottomrule
    \end{tabular}}
  \end{minipage}

  \vspace{-4pt}
\end{table}

\paragraph{\textbf{Metrics.}}
Follow established methods~\citep{dreambooth, wu2025less}, we consider 3 objective evaluation metrics across two aspects:  identity consistency, and text alignment.
\vspace{-0.2em}
\begin{itemize}[leftmargin=*, itemsep=0pt, topsep=0pt, parsep=0pt]
    \item Identity Consistency: We calculate the DINO-I Score~\citep{oquab2023dinov2} and CLIP-I~\citep{clip} Score between reference images and generated images to assess identity preservation.
    \item Text Alignment: We use the CLIP-T score~\citep{clip} to evaluate the model’s instruction-following ability.
\end{itemize}
\vspace{-0.2em}
We also incorporate subjective evaluation metrics: identity consistency, harmony, and text alignment to assess the performance of customization models.

\paragraph{\textbf{Baselines.}}
We compare our approach against several strong baselines, including the community FLUX.1 workflow (FLUX.1-Fill with FLUX.1-Redux)~\citep{FluxFill,FluxRedux}, state-of-the-art DiT-based open-source methods OminiCtrl~\citep{ominicontrol}, DreamO~\citep{mou2025dreamo}, and Insert Anything~\citep{song2025insert}, as well as the commercial system GPT-4o~\citep{gpt4o} (March 25, 2025).
Among them, FLUX.1 workflow and Insert Anything are primarily designed for position-aware customization, whereas OminiCtrl and DreamO target position-free customization. 
Beyond evaluating each method in its native setting, we also adapt the other baselines to complementary scenarios—feeding blank fill-in images to FLUX.1 workflow and Insert Anything to approximate position-free customization, and prompting OminiCtrl and DreamO with text descriptions of the identity embedded in the fill-in image scene to approximate position-aware customization. 
GPT-4o, in contrast, is a unified vision–language system. We therefore provide it with alternating image–text pairs and explicit instructions to perform each customization mode.
For completeness, and despite space constraints, we also include an evaluation of ACE++ in Appendix~Sec.~\ref{appendix:comparison_with_ace++}.

\begin{figure*}[t]
    \centering
    \includegraphics[width=.88\linewidth]{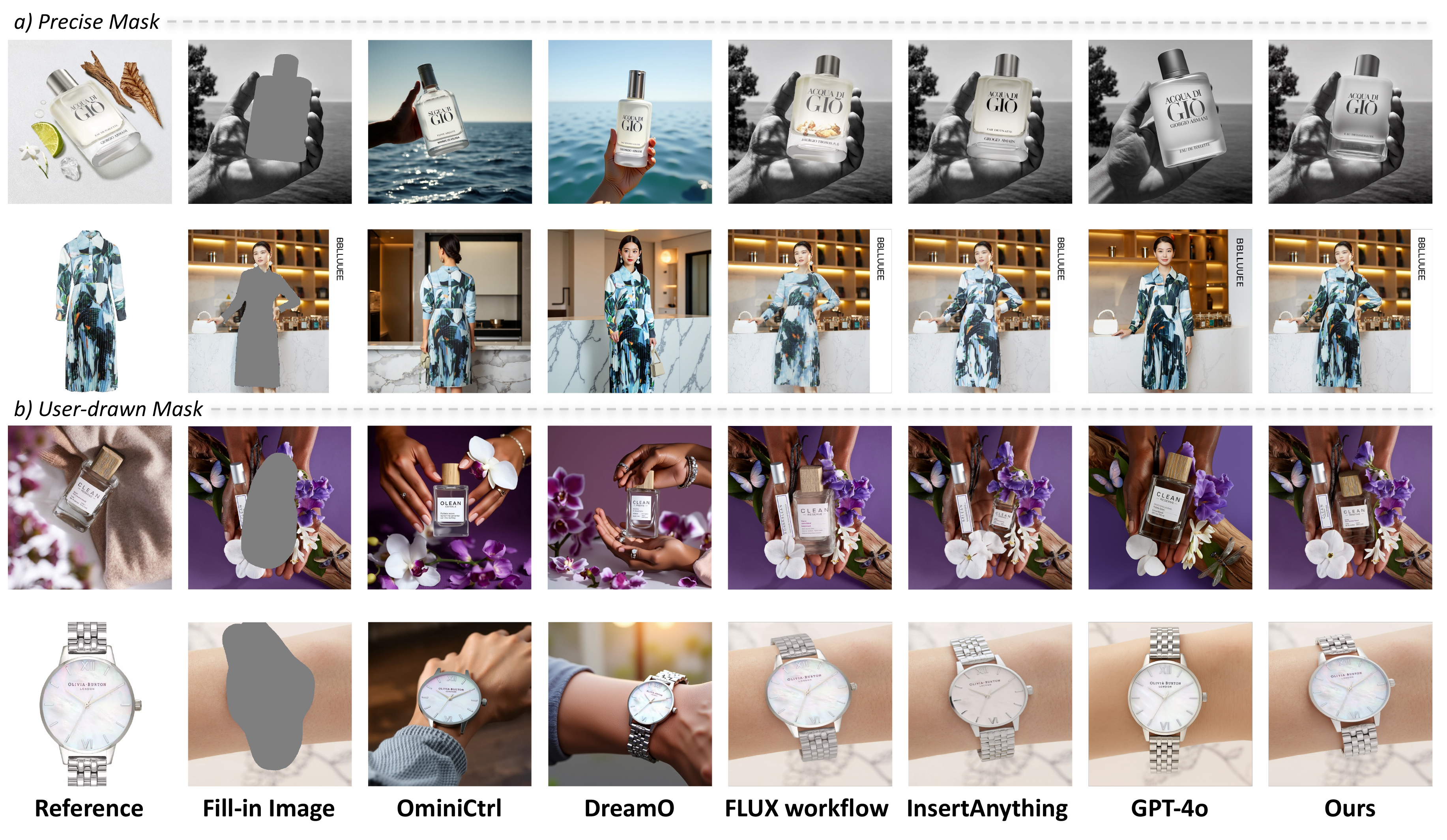}
\caption{\textbf{Qualitative comparison of position-aware customization under precise-mask and user-drawn-mask settings.} OminiCtrl and DreamO lack support for fill-in inputs. \OurMethod{} achieves high-quality customization with harmonious lighting, shadows, and perspectives.}
    \label{fig:comparisons-pos-aware}
\end{figure*}
\begin{figure*}[t]
    \centering
    \includegraphics[width=.88\linewidth]{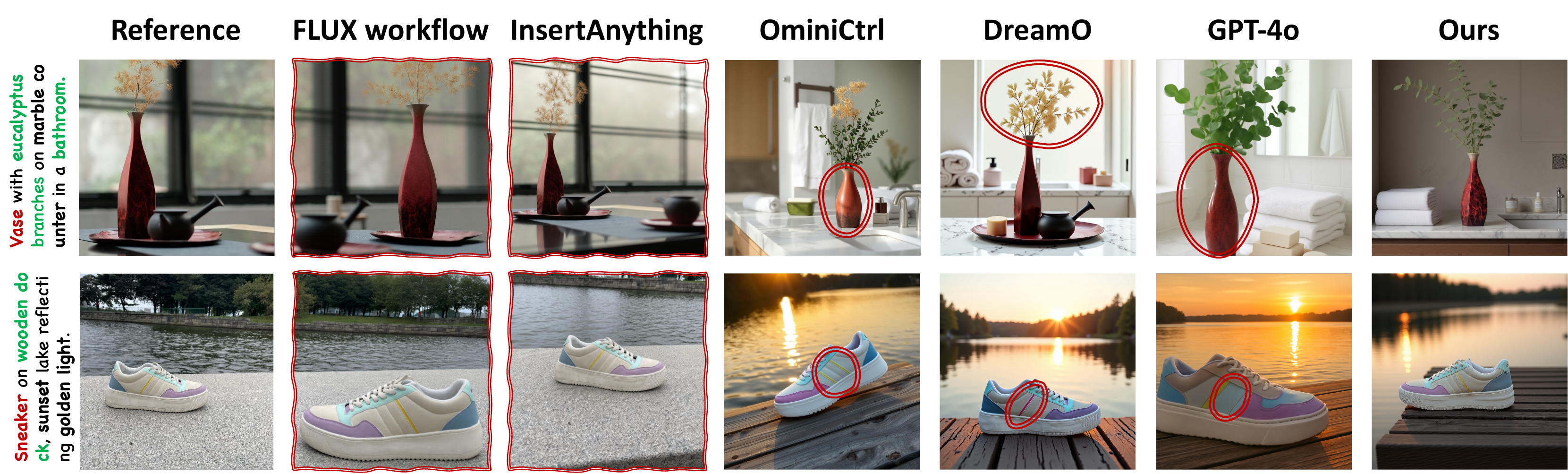}
    \caption{\textbf{Qualitative comparison on position-free customization.} \OurMethod{} achieves more realistic, coherent, and detailed customization. Red circles highlight incorrect regions or details.}
    \label{fig:comparisons-pos-free}
\end{figure*}

\subsection{Position-Aware Customization}
\paragraph{Quantitative Comparisons.}
Tab.~\ref{tab:objective_comparisons} reports quantitative results on \OurBench{} using both precise and user-drawn masks. 
\OurMethod{} achieves state-of-the-art identity consistency and text alignment, particularly under the more practical user-drawn mask setting (e.g., DINO-I 63.28 vs. 62.26). 
Although the adapted OminiCtrl, DreamO, and GPT-4o achieve reasonable scores, they essentially regenerate images rather than perform reference-based image filling (see the following paragraph). Despite being specifically designed for position-aware customization, FLUX.1 workflow and Insert Anything still underperform compared with our method.

\paragraph{Qualitative Comparisons.}
\label{sec:qualitative_pos_aware}
Fig.~\ref{fig:comparisons-pos-aware} presents qualitative comparisons on \OurBench{}.
OminiCtrl, DreamO, and GPT-4o tend to regenerate entire images rather than perform position-aware customization, for example in the precise-mask try-on case (second row) where the human’s face is completely altered.
FLUX.1 workflow and Insert Anything also produce noticeable artifacts and weaker identity preservation compared with our model. 
Moreover, under the user-drawn mask setting, our method generates content with harmonious size, shape, and appearance instead of merely filling the mask region.
Thanks to its unified in-context formulation, \OurMethod{} delivers position-aware customization with harmonious lighting, shadows, textures, and materials. More in Appendix Sec.~\ref{sec:more_visualization}.
More visual results are provided in Appendix Sec.~\ref{sec:more_visualization}.

\subsection{Position-Free Customization}
\paragraph{Quantitative Comparisons.}
FLUX.1 workflow and Insert Anything lack position-free customization capability and, even after adaptation, merely replicate the reference (see the following paragraph), so we exclude them. 
As shown in the DreamBench section of Tab.~\ref{tab:objective_comparisons}, OminiCtrl shows poor identity consistency (low DINO-I and CLIP-I), while DreamO and GPT-4o, though strong, still lag behind our approach.
Trained on a high-quality mix of real and synthetic data with a unified customization representation, our method achieves state-of-the-art performance across all metrics.

\paragraph{Qualitative Comparisons.}
Figure~\ref{fig:comparisons-pos-free} presents qualitative comparisons in the position-free setting. 
FLUX.1 workflow and Insert Anything fail to achieve true position-free customization, tending instead to replicate the reference identity image. 
OminiCtrl and DreamO produce results that are less realistic and less coherent than ours, while GPT-4o, despite strong instruction-following capabilities, sometimes loses fine-grained identity details. 
In contrast, \OurMethod{} consistently generates diverse, harmonious, and identity-consistent results.
More visual results are provided in Appendix Sec.~\ref{sec:more_visualization}.

\subsection{Human Evaluation}
We conducted a user study with 20 participants on 50 randomly selected samples from both position-aware and position-free subsets. For each sample, participants were asked to identify the best-performing model across three dimensions: identity consistency, harmony, and text alignment. As shown in Tab.~\ref{tab:human-ablation-side}(a), our method receives the highest human preference across all three dimensions compared with existing approaches. As FLUX.1 workflow and Insert Anything only take images as input, we exclude them from the rating of text alignment.

\subsection{Ablation Studies}
We present ablation studies of \OurMethod{} in Tab.~\ref{tab:human-ablation-side}(b), examining model architecture, training data sources, and training strategies. 
We first establish zero-shot performance as a baseline. 
We then validate several key design choices:  
\blackcircle{1} Without training the DiT image and text input layers (w/o IL), the model struggles to transfer the pre-trained diffusion prior to customization tasks, especially under user-drawn mask settings; 
\blackcircle{2} Training solely on synthetic data (w/o RD) weakens identity consistency and realism;  
\blackcircle{3} Omitting user-drawn mask data during training (w/o UM) substantially reduces performance on free-form masks;
\blackcircle{4} Removing Task-oriented Register tokens (w/o TR) or Boundary-aware Positional Embeddings (w/o PE) also degrades performance.  
Qualitative results in Sec.~\ref{sec:appendix_ablation_vis} confirm these findings: all ablated variants introduce artifacts or shape distortions, whereas our full model demonstrates superior flexibility and effectiveness.

\section{Conclusions and Limitations}
This paper presents \OurMethod{}, a flexible and effective framework for image customization. Our approach introduces four key contributions: (1) an in-context customization paradigm that unifies position-free and position-aware image customization; (2) a novel In-Context Multi-Modal Attention (ICMA) mechanism to adapt to different customization settings; (3) a high-quality identity-consistent dataset sourced primarily from real-world images; and (4) an evaluation benchmark with a balanced distribution of rigid and non-rigid customization tasks. Extensive experiments demonstrate that \OurMethod{} achieves state-of-the-art performance across multiple metrics.  

Despite these achievements, our method does not explicitly model viewpoint, lighting, geometry, or other 3D scene properties, which we plan to address in future work. We also provide an initial exploration of multi-reference customization in Appendix~\ref{appendix:multi-ref}.

\bibliography{main}

\begin{thebibliography}{53}
\providecommand{\natexlab}[1]{#1}
\providecommand{\url}[1]{\texttt{#1}}
\expandafter\ifx\csname urlstyle\endcsname\relax
  \providecommand{\doi}[1]{doi: #1}\else
  \providecommand{\doi}{doi: \begingroup \urlstyle{rm}\Url}\fi

\bibitem[Avrahami et~al.(2023)Avrahami, Aberman, Fried, Cohen-Or, and Lischinski]{avrahami2023break}
Omri Avrahami, Kfir Aberman, Ohad Fried, Daniel Cohen-Or, and Dani Lischinski.
\newblock Break-a-scene: Extracting multiple concepts from a single image.
\newblock In \emph{SIGGRAPH Asia 2023 Conference Papers}, pp.\  1--12, 2023.

\bibitem[Bai et~al.(2025)Bai, Chen, Liu, Wang, Ge, Song, Dang, Wang, Wang, Tang, et~al.]{bai2025qwen2}
Shuai Bai, Keqin Chen, Xuejing Liu, Jialin Wang, Wenbin Ge, Sibo Song, Kai Dang, Peng Wang, Shijie Wang, Jun Tang, et~al.
\newblock Qwen2. 5-vl technical report.
\newblock \emph{arXiv preprint arXiv:2502.13923}, 2025.

\bibitem[Chen et~al.(2024{\natexlab{a}})Chen, Feng, Chen, Wang, Zhang, Liu, Shen, and Zhao]{chen2024zero}
Xi~Chen, Yutong Feng, Mengting Chen, Yiyang Wang, Shilong Zhang, Yu~Liu, Yujun Shen, and Hengshuang Zhao.
\newblock Zero-shot image editing with reference imitation.
\newblock \emph{Advances in Neural Information Processing Systems}, 37:\penalty0 84010--84032, 2024{\natexlab{a}}.

\bibitem[Chen et~al.(2024{\natexlab{b}})Chen, Huang, Liu, Shen, Zhao, and Zhao]{chen2024anydoor}
Xi~Chen, Lianghua Huang, Yu~Liu, Yujun Shen, Deli Zhao, and Hengshuang Zhao.
\newblock Anydoor: Zero-shot object-level image customization.
\newblock In \emph{Proceedings of the IEEE/CVF conference on computer vision and pattern recognition}, pp.\  6593--6602, 2024{\natexlab{b}}.

\bibitem[Chen et~al.(2024{\natexlab{c}})Chen, Zhang, Zhang, Zhou, Kim, Liu, Li, Zhang, Zhao, Wang, et~al.]{unireal}
Xi~Chen, Zhifei Zhang, He~Zhang, Yuqian Zhou, Soo~Ye Kim, Qing Liu, Yijun Li, Jianming Zhang, Nanxuan Zhao, Yilin Wang, et~al.
\newblock Unireal: Universal image generation and editing via learning real-world dynamics.
\newblock \emph{arXiv preprint arXiv:2412.07774}, 2024{\natexlab{c}}.

\bibitem[Dhariwal \& Nichol(2021)Dhariwal and Nichol]{dhariwal2021diffusion}
Prafulla Dhariwal and Alexander Nichol.
\newblock Diffusion models beat gans on image synthesis.
\newblock \emph{Advances in neural information processing systems}, 34:\penalty0 8780--8794, 2021.

\bibitem[Esser et~al.(2024{\natexlab{a}})Esser, Kulal, Blattmann, Entezari, M{\"u}ller, Saini, Levi, Lorenz, Sauer, Boesel, et~al.]{esser2024scaling}
Patrick Esser, Sumith Kulal, Andreas Blattmann, Rahim Entezari, Jonas M{\"u}ller, Harry Saini, Yam Levi, Dominik Lorenz, Axel Sauer, Frederic Boesel, et~al.
\newblock Scaling rectified flow transformers for high-resolution image synthesis.
\newblock In \emph{Forty-first international conference on machine learning}, 2024{\natexlab{a}}.

\bibitem[Esser et~al.(2024{\natexlab{b}})Esser, Kulal, Blattmann, Entezari, M{\"u}ller, Saini, Levi, Lorenz, Sauer, Boesel, et~al.]{sd3}
Patrick Esser, Sumith Kulal, Andreas Blattmann, Rahim Entezari, Jonas M{\"u}ller, Harry Saini, Yam Levi, Dominik Lorenz, Axel Sauer, Frederic Boesel, et~al.
\newblock Scaling rectified flow transformers for high-resolution image synthesis.
\newblock In \emph{Forty-first international conference on machine learning}, 2024{\natexlab{b}}.

\bibitem[Gal et~al.(2022)Gal, Alaluf, Atzmon, Patashnik, Bermano, Chechik, and Cohen-Or]{gal2022image}
Rinon Gal, Yuval Alaluf, Yuval Atzmon, Or~Patashnik, Amit~H Bermano, Gal Chechik, and Daniel Cohen-Or.
\newblock An image is worth one word: Personalizing text-to-image generation using textual inversion.
\newblock \emph{arXiv preprint arXiv:2208.01618}, 2022.

\bibitem[Goodfellow et~al.(2020)Goodfellow, Pouget-Abadie, Mirza, Xu, Warde-Farley, Ozair, Courville, and Bengio]{goodfellow2020generative}
Ian Goodfellow, Jean Pouget-Abadie, Mehdi Mirza, Bing Xu, David Warde-Farley, Sherjil Ozair, Aaron Courville, and Yoshua Bengio.
\newblock Generative adversarial networks.
\newblock \emph{Communications of the ACM}, 63\penalty0 (11):\penalty0 139--144, 2020.

\bibitem[Guo et~al.(2025)Guo, Zeng, Song, Zhang, Zhang, and Liu]{guo2025any2anytryon}
Hailong Guo, Bohan Zeng, Yiren Song, Wentao Zhang, Chuang Zhang, and Jiaming Liu.
\newblock Any2anytryon: Leveraging adaptive position embeddings for versatile virtual clothing tasks.
\newblock \emph{arXiv preprint arXiv:2501.15891}, 2025.

\bibitem[Ho et~al.(2020)Ho, Jain, and Abbeel]{ho2020denoising}
Jonathan Ho, Ajay Jain, and Pieter Abbeel.
\newblock Denoising diffusion probabilistic models.
\newblock \emph{Advances in neural information processing systems}, 33:\penalty0 6840--6851, 2020.

\bibitem[Hu et~al.(2022)Hu, Shen, Wallis, Allen-Zhu, Li, Wang, Wang, Chen, et~al.]{hu2022lora}
Edward~J Hu, Yelong Shen, Phillip Wallis, Zeyuan Allen-Zhu, Yuanzhi Li, Shean Wang, Lu~Wang, Weizhu Chen, et~al.
\newblock Lora: Low-rank adaptation of large language models.
\newblock \emph{ICLR}, 1\penalty0 (2):\penalty0 3, 2022.

\bibitem[Hurst et~al.(2024{\natexlab{a}})Hurst, Lerer, Goucher, Perelman, Ramesh, Clark, Ostrow, Welihinda, Hayes, Radford, et~al.]{gpt4o}
Aaron Hurst, Adam Lerer, Adam~P Goucher, Adam Perelman, Aditya Ramesh, Aidan Clark, AJ~Ostrow, Akila Welihinda, Alan Hayes, Alec Radford, et~al.
\newblock Gpt-4o system card.
\newblock \emph{arXiv preprint arXiv:2410.21276}, 2024{\natexlab{a}}.

\bibitem[Hurst et~al.(2024{\natexlab{b}})Hurst, Lerer, Goucher, Perelman, Ramesh, Clark, Ostrow, Welihinda, Hayes, Radford, et~al.]{hurst2024gpt}
Aaron Hurst, Adam Lerer, Adam~P Goucher, Adam Perelman, Aditya Ramesh, Aidan Clark, AJ~Ostrow, Akila Welihinda, Alan Hayes, Alec Radford, et~al.
\newblock Gpt-4o system card.
\newblock \emph{arXiv preprint arXiv:2410.21276}, 2024{\natexlab{b}}.

\bibitem[Kingma et~al.(2013{\natexlab{a}})Kingma, Welling, et~al.]{kingma2013auto}
Diederik~P Kingma, Max Welling, et~al.
\newblock Auto-encoding variational bayes, 2013{\natexlab{a}}.

\bibitem[Kingma et~al.(2013{\natexlab{b}})Kingma, Welling, et~al.]{vae}
Diederik~P Kingma, Max Welling, et~al.
\newblock Auto-encoding variational bayes, 2013{\natexlab{b}}.

\bibitem[Kirillov et~al.(2023)Kirillov, Mintun, Ravi, Mao, Rolland, Gustafson, Xiao, Whitehead, Berg, Lo, et~al.]{kirillov2023segment}
Alexander Kirillov, Eric Mintun, Nikhila Ravi, Hanzi Mao, Chloe Rolland, Laura Gustafson, Tete Xiao, Spencer Whitehead, Alexander~C Berg, Wan-Yen Lo, et~al.
\newblock Segment anything.
\newblock In \emph{Proceedings of the IEEE/CVF International Conference on Computer Vision (ICCV)}, October 2023.

\bibitem[Kumari et~al.(2025)Kumari, Yin, Zhu, Misra, and Azadi]{kumari2025generating}
Nupur Kumari, Xi~Yin, Jun-Yan Zhu, Ishan Misra, and Samaneh Azadi.
\newblock Generating multi-image synthetic data for text-to-image customization.
\newblock \emph{arXiv preprint arXiv:2502.01720}, 2025.

\bibitem[Labs(2024{\natexlab{a}})]{Flux}
Black~Forest Labs.
\newblock Flux.
\newblock \url{https://github.com/black-forest-labs/flux}, 2024{\natexlab{a}}.

\bibitem[Labs(2024{\natexlab{b}})]{FluxFill}
Black~Forest Labs.
\newblock Flux.1-fill-dev.
\newblock \url{https://huggingface.co/black-forest-labs/FLUX.1-Fill-dev}, 2024{\natexlab{b}}.

\bibitem[Labs(2024{\natexlab{c}})]{FluxRedux}
Black~Forest Labs.
\newblock Flux.1-redux-dev.
\newblock \url{https://huggingface.co/black-forest-labs/FLUX.1-Redux-dev/}, 2024{\natexlab{c}}.

\bibitem[Labs et~al.(2025)Labs, Batifol, Blattmann, Boesel, Consul, Diagne, Dockhorn, English, English, Esser, et~al.]{labs2025flux}
Black~Forest Labs, Stephen Batifol, Andreas Blattmann, Frederic Boesel, Saksham Consul, Cyril Diagne, Tim Dockhorn, Jack English, Zion English, Patrick Esser, et~al.
\newblock Flux. 1 kontext: Flow matching for in-context image generation and editing in latent space.
\newblock \emph{arXiv preprint arXiv:2506.15742}, 2025.

\bibitem[Li et~al.(2025)Li, Du, Yan, Zhuo, Li, Gao, Ma, and Cheng]{li2025visualcloze}
Zhong-Yu Li, Ruoyi Du, Juncheng Yan, Le~Zhuo, Zhen Li, Peng Gao, Zhanyu Ma, and Ming-Ming Cheng.
\newblock Visualcloze: A universal image generation framework via visual in-context learning.
\newblock \emph{arXiv preprint arXiv:2504.07960}, 2025.

\bibitem[Lipman et~al.(2023)Lipman, Chen, Ben-Hamu, Nickel, and Le]{lipman2023flow}
Yaron Lipman, Ricky~TQ Chen, Heli Ben-Hamu, Maximilian Nickel, and Matt Le.
\newblock Flow matching for generative modeling.
\newblock In \emph{11th International Conference on Learning Representations, ICLR 2023}, 2023.

\bibitem[Liu et~al.(2022)Liu, Gong, and Liu]{rectified_flow}
Xingchao Liu, Chengyue Gong, and Qiang Liu.
\newblock Flow straight and fast: Learning to generate and transfer data with rectified flow.
\newblock \emph{arXiv preprint arXiv:2209.03003}, 2022.

\bibitem[Loshchilov \& Hutter(2017)Loshchilov and Hutter]{adam}
Ilya Loshchilov and Frank Hutter.
\newblock Decoupled weight decay regularization.
\newblock \emph{arXiv preprint arXiv:1711.05101}, 2017.

\bibitem[Mao et~al.(2025)Mao, Zhang, Pan, Jiang, Han, Liu, and Zhou]{mao2025ace++}
Chaojie Mao, Jingfeng Zhang, Yulin Pan, Zeyinzi Jiang, Zhen Han, Yu~Liu, and Jingren Zhou.
\newblock Ace++: Instruction-based image creation and editing via context-aware content filling.
\newblock \emph{arXiv preprint arXiv:2501.02487}, 2025.

\bibitem[Mou et~al.(2025)Mou, Wu, Wu, Guo, Zhang, Cheng, Luo, Ding, Zhang, Li, et~al.]{mou2025dreamo}
Chong Mou, Yanze Wu, Wenxu Wu, Zinan Guo, Pengze Zhang, Yufeng Cheng, Yiming Luo, Fei Ding, Shiwen Zhang, Xinghui Li, et~al.
\newblock Dreamo: A unified framework for image customization.
\newblock \emph{arXiv preprint arXiv:2504.16915}, 2025.

\bibitem[Nichol et~al.(2021)Nichol, Dhariwal, Ramesh, Shyam, Mishkin, McGrew, Sutskever, and Chen]{nichol2021glide}
Alex Nichol, Prafulla Dhariwal, Aditya Ramesh, Pranav Shyam, Pamela Mishkin, Bob McGrew, Ilya Sutskever, and Mark Chen.
\newblock Glide: Towards photorealistic image generation and editing with text-guided diffusion models.
\newblock \emph{arXiv preprint arXiv:2112.10741}, 2021.

\bibitem[Oquab et~al.(2023)Oquab, Darcet, Moutakanni, Vo, Szafraniec, Khalidov, Fernandez, Haziza, Massa, El-Nouby, et~al.]{oquab2023dinov2}
Maxime Oquab, Timoth{\'e}e Darcet, Th{\'e}o Moutakanni, Huy Vo, Marc Szafraniec, Vasil Khalidov, Pierre Fernandez, Daniel Haziza, Francisco Massa, Alaaeldin El-Nouby, et~al.
\newblock Dinov2: Learning robust visual features without supervision.
\newblock \emph{arXiv preprint arXiv:2304.07193}, 2023.

\bibitem[Peebles \& Xie(2023{\natexlab{a}})Peebles and Xie]{dit}
William Peebles and Saining Xie.
\newblock Scalable diffusion models with transformers.
\newblock In \emph{Proceedings of the IEEE/CVF international conference on computer vision}, pp.\  4195--4205, 2023{\natexlab{a}}.

\bibitem[Peebles \& Xie(2023{\natexlab{b}})Peebles and Xie]{peebles2023scalable}
William Peebles and Saining Xie.
\newblock Scalable diffusion models with transformers.
\newblock In \emph{Proceedings of the IEEE/CVF international conference on computer vision}, pp.\  4195--4205, 2023{\natexlab{b}}.

\bibitem[Radford et~al.(2021)Radford, Kim, Hallacy, Ramesh, Goh, Agarwal, Sastry, Askell, Mishkin, Clark, et~al.]{clip}
Alec Radford, Jong~Wook Kim, Chris Hallacy, Aditya Ramesh, Gabriel Goh, Sandhini Agarwal, Girish Sastry, Amanda Askell, Pamela Mishkin, Jack Clark, et~al.
\newblock Learning transferable visual models from natural language supervision.
\newblock In \emph{International conference on machine learning}, pp.\  8748--8763, 2021.

\bibitem[Raffel et~al.(2020)Raffel, Shazeer, Roberts, Lee, Narang, Matena, Zhou, Li, and Liu]{raffel2020exploring}
Colin Raffel, Noam Shazeer, Adam Roberts, Katherine Lee, Sharan Narang, Michael Matena, Yanqi Zhou, Wei Li, and Peter~J Liu.
\newblock Exploring the limits of transfer learning with a unified text-to-text transformer.
\newblock \emph{Journal of machine learning research}, 21\penalty0 (140):\penalty0 1--67, 2020.

\bibitem[Ramesh et~al.(2022)Ramesh, Dhariwal, Nichol, Chu, and Chen]{ramesh2022hierarchical}
Aditya Ramesh, Prafulla Dhariwal, Alex Nichol, Casey Chu, and Mark Chen.
\newblock Hierarchical text-conditional image generation with clip latents.
\newblock \emph{arXiv preprint arXiv:2204.06125}, 1\penalty0 (2):\penalty0 3, 2022.

\bibitem[Ren et~al.(2024)Ren, Liu, Zeng, Lin, Li, Cao, Chen, Huang, Chen, Yan, et~al.]{ren2024grounded}
Tianhe Ren, Shilong Liu, Ailing Zeng, Jing Lin, Kunchang Li, He~Cao, Jiayu Chen, Xinyu Huang, Yukang Chen, Feng Yan, et~al.
\newblock Grounded sam: Assembling open-world models for diverse visual tasks.
\newblock \emph{arXiv preprint arXiv:2401.14159}, 2024.

\bibitem[Rombach et~al.(2022)Rombach, Blattmann, Lorenz, Esser, and Ommer]{rombach2022high}
Robin Rombach, Andreas Blattmann, Dominik Lorenz, Patrick Esser, and Bj{\"o}rn Ommer.
\newblock High-resolution image synthesis with latent diffusion models.
\newblock In \emph{Proceedings of the IEEE/CVF conference on computer vision and pattern recognition}, pp.\  10684--10695, 2022.

\bibitem[Ronneberger et~al.(2015)Ronneberger, Fischer, and Brox]{ronneberger2015u}
Olaf Ronneberger, Philipp Fischer, and Thomas Brox.
\newblock U-net: Convolutional networks for biomedical image segmentation.
\newblock In \emph{Medical image computing and computer-assisted intervention--MICCAI 2015: 18th international conference, Munich, Germany, October 5-9, 2015, proceedings, part III 18}, pp.\  234--241. Springer, 2015.

\bibitem[Ruiz et~al.(2023{\natexlab{a}})Ruiz, Li, Jampani, Pritch, Rubinstein, and Aberman]{dreambooth}
Nataniel Ruiz, Yuanzhen Li, Varun Jampani, Yael Pritch, Michael Rubinstein, and Kfir Aberman.
\newblock Dreambooth: Fine tuning text-to-image diffusion models for subject-driven generation.
\newblock In \emph{Proceedings of the IEEE/CVF conference on computer vision and pattern recognition}, pp.\  22500--22510, 2023{\natexlab{a}}.

\bibitem[Ruiz et~al.(2023{\natexlab{b}})Ruiz, Li, Jampani, Pritch, Rubinstein, and Aberman]{ruiz2023dreambooth}
Nataniel Ruiz, Yuanzhen Li, Varun Jampani, Yael Pritch, Michael Rubinstein, and Kfir Aberman.
\newblock Dreambooth: Fine tuning text-to-image diffusion models for subject-driven generation.
\newblock In \emph{Proceedings of the IEEE/CVF conference on computer vision and pattern recognition}, pp.\  22500--22510, 2023{\natexlab{b}}.

\bibitem[Sohl-Dickstein et~al.(2015)Sohl-Dickstein, Weiss, Maheswaranathan, and Ganguli]{sohl2015deep}
Jascha Sohl-Dickstein, Eric Weiss, Niru Maheswaranathan, and Surya Ganguli.
\newblock Deep unsupervised learning using nonequilibrium thermodynamics.
\newblock In \emph{International conference on machine learning}, pp.\  2256--2265. pmlr, 2015.

\bibitem[Song et~al.(2025)Song, Jiang, Yang, Quan, and Yang]{song2025insert}
Wensong Song, Hong Jiang, Zongxing Yang, Ruijie Quan, and Yi~Yang.
\newblock Insert anything: Image insertion via in-context editing in dit.
\newblock \emph{arXiv preprint arXiv:2504.15009}, 2025.

\bibitem[Su et~al.(2024)Su, Ahmed, Lu, Pan, Bo, and Liu]{su2024roformer}
Jianlin Su, Murtadha Ahmed, Yu~Lu, Shengfeng Pan, Wen Bo, and Yunfeng Liu.
\newblock Roformer: Enhanced transformer with rotary position embedding.
\newblock \emph{Neurocomputing}, 568:\penalty0 127063, 2024.

\bibitem[Tan et~al.(2024)Tan, Liu, Yang, Xue, and Wang]{ominicontrol}
Zhenxiong Tan, Songhua Liu, Xingyi Yang, Qiaochu Xue, and Xinchao Wang.
\newblock Ominicontrol: Minimal and universal control for diffusion transformer.
\newblock \emph{arXiv preprint arXiv:2411.15098}, 2024.

\bibitem[Team et~al.(2023)Team, Anil, Borgeaud, Alayrac, Yu, Soricut, Schalkwyk, Dai, Hauth, Millican, et~al.]{team2023gemini}
Gemini Team, Rohan Anil, Sebastian Borgeaud, Jean-Baptiste Alayrac, Jiahui Yu, Radu Soricut, Johan Schalkwyk, Andrew~M Dai, Anja Hauth, Katie Millican, et~al.
\newblock Gemini: a family of highly capable multimodal models.
\newblock \emph{arXiv preprint arXiv:2312.11805}, 2023.

\bibitem[Tewel et~al.(2024)Tewel, Kaduri, Gal, Kasten, Wolf, Chechik, and Atzmon]{tewel2024training}
Yoad Tewel, Omri Kaduri, Rinon Gal, Yoni Kasten, Lior Wolf, Gal Chechik, and Yuval Atzmon.
\newblock Training-free consistent text-to-image generation.
\newblock \emph{ACM Transactions on Graphics (TOG)}, 43\penalty0 (4):\penalty0 1--18, 2024.

\bibitem[Vaswani et~al.(2017)Vaswani, Shazeer, Parmar, Uszkoreit, Jones, Gomez, Kaiser, and Polosukhin]{vaswani2017attention}
Ashish Vaswani, Noam Shazeer, Niki Parmar, Jakob Uszkoreit, Llion Jones, Aidan~N Gomez, {\L}ukasz Kaiser, and Illia Polosukhin.
\newblock Attention is all you need.
\newblock \emph{Advances in neural information processing systems}, 30, 2017.

\bibitem[Wang et~al.(2024)Wang, Guo, Liu, Li, Zhao, Tang, Hu, Tang, and Li]{wang2024stablegarment}
Rui Wang, Hailong Guo, Jiaming Liu, Huaxia Li, Haibo Zhao, Xu~Tang, Yao Hu, Hao Tang, and Peipei Li.
\newblock Stablegarment: Garment-centric generation via stable diffusion.
\newblock \emph{arXiv preprint arXiv:2403.10783}, 2024.

\bibitem[Wu et~al.(2025)Wu, Huang, Wu, Cheng, Ding, and He]{wu2025less}
Shaojin Wu, Mengqi Huang, Wenxu Wu, Yufeng Cheng, Fei Ding, and Qian He.
\newblock Less-to-more generalization: Unlocking more controllability by in-context generation.
\newblock \emph{arXiv preprint arXiv:2504.02160}, 2025.

\bibitem[Xu et~al.(2025)Xu, Gu, Chen, and Chen]{xu2025ootdiffusion}
Yuhao Xu, Tao Gu, Weifeng Chen, and Arlene Chen.
\newblock Ootdiffusion: Outfitting fusion based latent diffusion for controllable virtual try-on.
\newblock In \emph{Proceedings of the AAAI Conference on Artificial Intelligence}, volume~39, pp.\  8996--9004, 2025.

\bibitem[Ye et~al.(2023{\natexlab{a}})Ye, Zhang, Liu, Han, and Yang]{ipadapter}
Hu~Ye, Jun Zhang, Sibo Liu, Xiao Han, and Wei Yang.
\newblock Ip-adapter: Text compatible image prompt adapter for text-to-image diffusion models.
\newblock \emph{arXiv preprint arXiv:2308.06721}, 2023{\natexlab{a}}.

\bibitem[Ye et~al.(2023{\natexlab{b}})Ye, Zhang, Liu, Han, and Yang]{ye2023ip}
Hu~Ye, Jun Zhang, Sibo Liu, Xiao Han, and Wei Yang.
\newblock Ip-adapter: Text compatible image prompt adapter for text-to-image diffusion models.
\newblock \emph{arXiv preprint arXiv:2308.06721}, 2023{\natexlab{b}}.

\end{thebibliography}
\bibliographystyle{main}

\newpage
\appendix

\part*{Appendix}
\addcontentsline{toc}{part}{Appendix} 
\etocsetnexttocdepth{subsection}

\localtableofcontents
\clearpage

\section{Related Work}

\subsection{Image Diffusion Models}
Recent advances in diffusion models~\citep{sohl2015deep, ho2020denoising} have set new benchmarks in image synthesis, outperforming traditional generative models such as Variational Autoencoders (VAE)~\citep{kingma2013auto} and Generative Adversarial Networks (GANs)~\citep{goodfellow2020generative} by a significant margin. Consequently, many state-of-the-art text-to-image methods~\citep{dhariwal2021diffusion, ho2020denoising, nichol2021glide, ramesh2022hierarchical} have adopted diffusion models as their core generation framework. Early approaches employed a U-Net~\citep{ronneberger2015u} architecture with cross-attention for text-to-image generation, achieving competitive performance and efficiency. Notably, the open-sourcing of Stable Diffusion~\citep{rombach2022high} has been a major catalyst for the growth of image synthesis research. More recently, diffusion transformer models, such as SD3~\citep{sd3} and FLUX~\citep{Flux}, have further advanced the field by integrating transformer architectures~\citep{vaswani2017attention} with diffusion models, yielding even higher performance. These models have since been widely applied in various downstream tasks, including depth estimation, image editing, and others.

\subsection{Image Customization}
Image customization is typically accomplished by integrating additional control signals from reference images into text-to-image foundation models. One line of work~\citep{wu2025less, li2025visualcloze, gpt4o, mou2025dreamo, ominicontrol, unireal} focuses on position-free customization, directly generating identity-consistent images based on input reference images and text, as seen in GPT-4o~\citep{gpt4o}, DreamO~\citep{mou2025dreamo}, and OminiControl~\citep{ominicontrol}. However, these methods struggle with position-aware customization, particularly when a masked source image is provided, as they cannot preserve the unedited regions. In contrast, methods like Insert Anything~\citep{song2025insert} and the FLUX.1-Fill-Redux workflow~\citep{FluxFill} specialize in position-aware customization, inserting subjects into masked source images, but lack the capability for position-free customization.
Concurrent works such as ACE++~\citep{mao2025ace++} and FLUX.1 Kontext~\citep{labs2025flux} share similar ideas with our approach, yet differ in innovative technical details.
In this work, we propose a flexible framework that can address both position-aware customization and position-free customization. We also propose a data curation pipeline to collect high-quality real image data from different product images. Benefiting from this framework and high-quality data, our model achieves highly identity consistent customization, which can be used in real production.

\section{Ablation Visualization}
\label{sec:appendix_ablation_vis}
\begin{figure*}[ht]
    \centering
    \includegraphics[width=1.\linewidth]{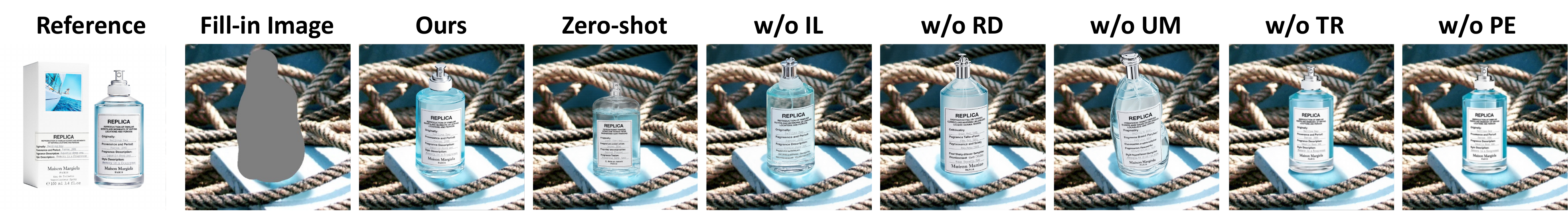}
    \caption{\textbf{Ablation Visualization.} Qualitative results show that our model preserves identity consistency while enabling harmonious customization. Abbreviations are as follows: Zero-shot = zero-shot inference without fine-tuning; 
    w/o IL = without training Input Layers; 
    w/o RD = without using Real Data for training; 
    w/o UM = without using User-drawn Mask for training;
    w/o TR = without Task-oriented Register tokens; 
    w/o PE = without Boundary-aware Positional Embeddings.}
    \label{fig:ablations}
\end{figure*}

\begin{figure*}[ht]
    \centering
    \includegraphics[width=1.\linewidth]{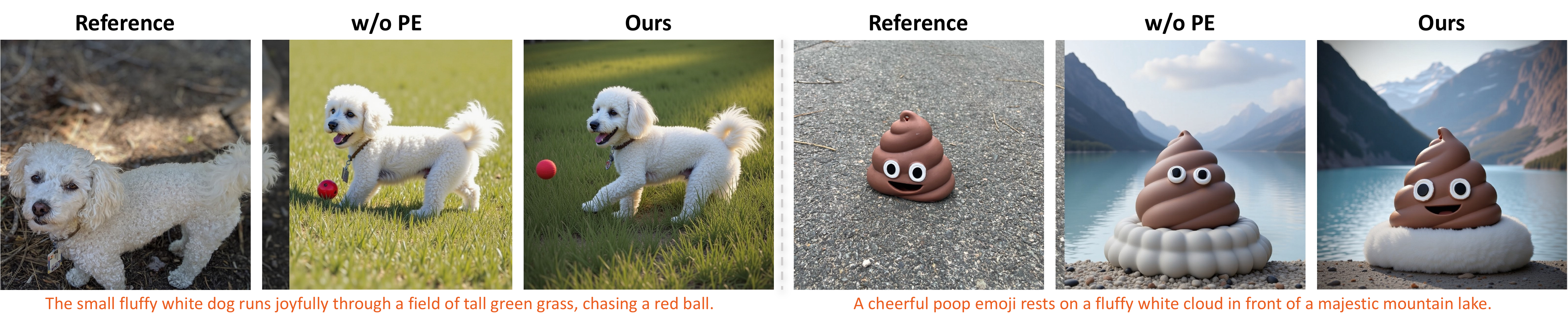}
   \caption{\textbf{Effect of Boundary-aware Positional Embeddings.}  Without Boundary-aware Positional Embeddings (PE), position-free customization can produce blurred or ambiguous boundaries between the reference and generated content. Incorporating these embeddings sharpens boundaries.}
    \label{fig:ablation_boundary_pos_embeddings}
\end{figure*}

As shown in Fig.~\ref{fig:ablations}, we visualize the ablation cases.  
Other variants either fail to preserve the reference identity or produce incoherent, distorted customization results.  
In contrast, our full model preserves identity while naturally integrating it into the scene, yielding harmonious lighting and perspective.  
We also observe that in position-free customization, performing flow matching on both the reference and output images can blur their boundaries—an issue alleviated by incorporating Boundary-aware Positional Embeddings (see Fig.~\ref{fig:ablation_boundary_pos_embeddings}).


\section{Automated Captioning for Data}
\label{sec:automated_caption}
\begin{figure*}[t]
    \centering
    \includegraphics[width=\linewidth]{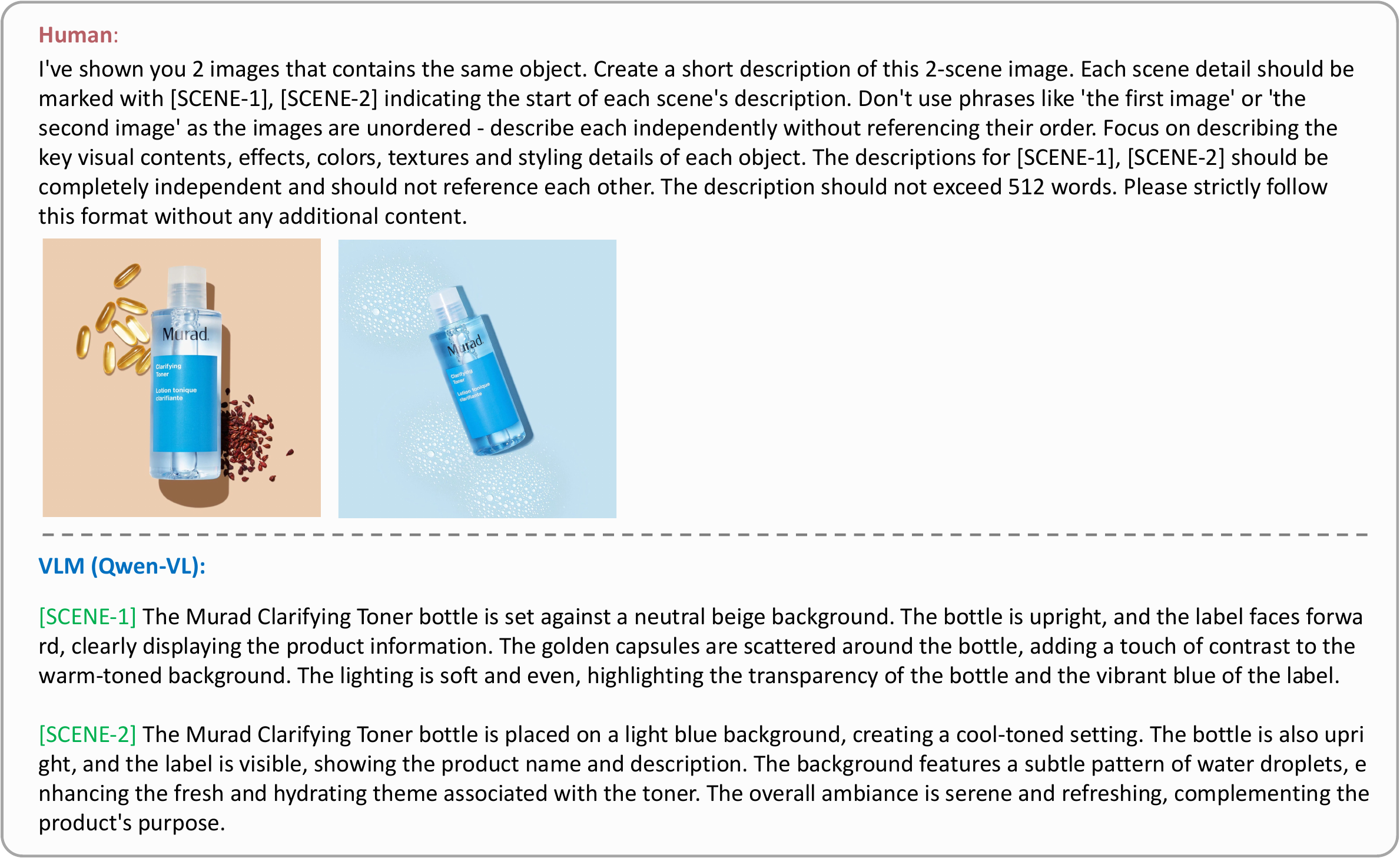}
    \caption{\textbf{Example of automated text prompt annotation.}  
    A concatenated pair of identity-consistent images is fed into Qwen-VL2.5~\citep{bai2025qwen2} with custom-designed instructions to generate corresponding captions for our data.}
    \label{fig:data_caption}
\end{figure*}

We use Qwen-VL2.5~\citep{bai2025qwen2} to automatically generate text annotations for our data. Specifically, each concatenated pair of identity-consistent images is fed into Qwen-VL2.5 with custom-designed instructions to generate captions, as illustrated in Fig.~\ref{fig:data_caption}.
\section{Automated Captioning for Benchmark}
\label{sec:automated_caption_for_benchmark}
\begin{figure*}[t]
    \centering
    \includegraphics[width=\linewidth]{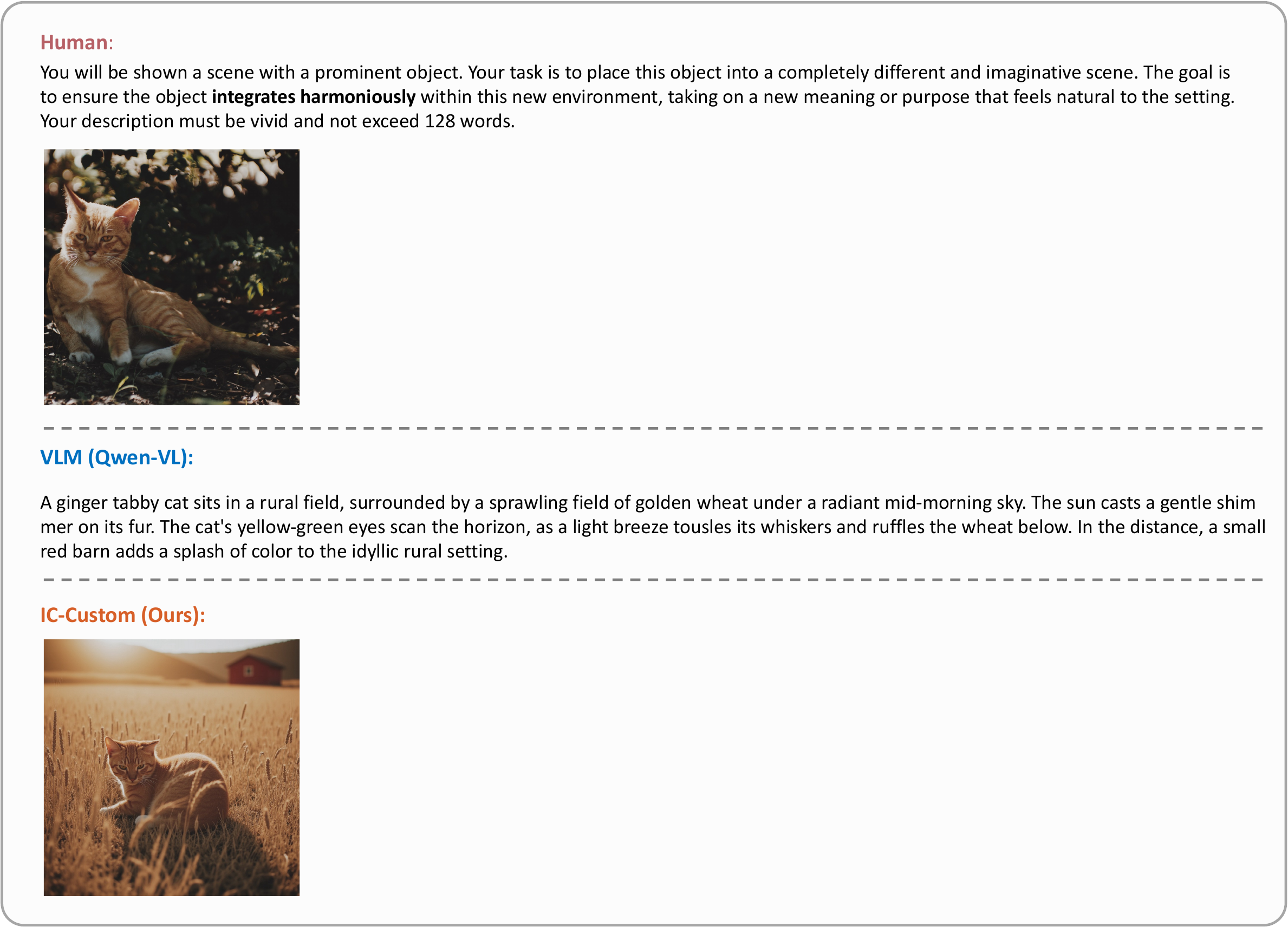}
    \caption{\textbf{Example of DreamBench captioning and generated output.}  
    We illustrate our prompting process for DreamBench, where a reference image and custom instructions are provided to a vision–language model to generate creative, identity-consistent captions. The figure also shows an example image generated by our method using the curated reference and caption.}
    
    \label{fig:dreambench_example}
\end{figure*}

Our benchmark consists of two parts: \OurBench{} for evaluating position-aware customization and DreamBench~\citep{dreambooth} for evaluating position-free customization. For \OurBench{}, we apply the captioning approach described in Sec.~\ref{sec:automated_caption} to generate input captions. For DreamBench, which targets position-free customization, we provide the reference image together with prompts designed to elicit creative yet identity-consistent outputs; an example of this prompting strategy is shown in Fig.~\ref{fig:dreambench_example}.

\section{Comparison with ACE++}
\label{appendix:comparison_with_ace++}
\begin{figure}[ht]
    \centering
    \includegraphics[width=1.\linewidth]{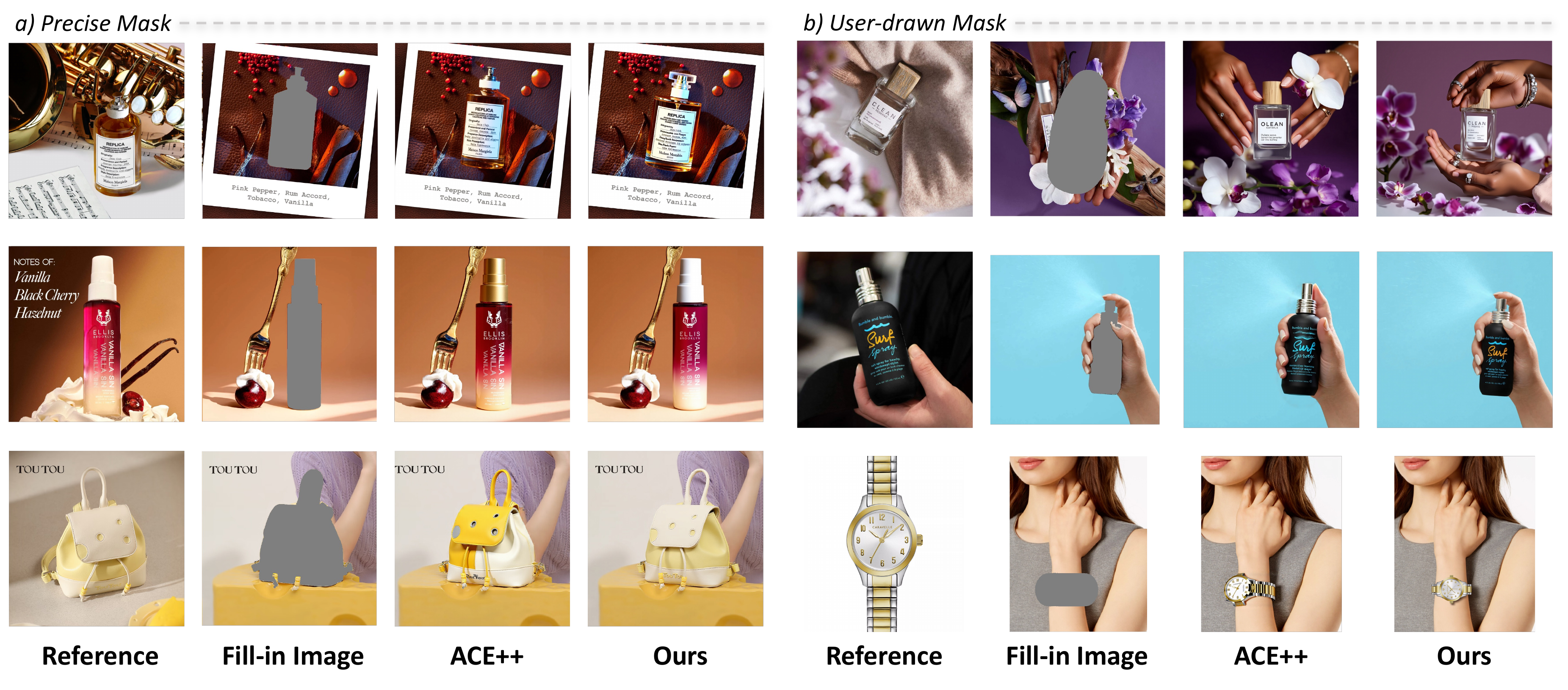}
\caption{\textbf{Qualitative comparison with ACE++~\citep{mao2025ace++}. Our method produces more identity-consistent and harmonious customization results.}  
We compare our unified framework with ACE++ on \OurBench{}.}
    \label{fig:comparisons-ace++}
\end{figure}

ACE++~\citep{mao2025ace++} is a concurrent work proposing the Long-context Condition Unit (LCU), conceptually similar to our in-context diptych. However, ACE++ focuses on four separate domain-specific tasks and trains distinct LoRA adapters for each, rather than a unified model handling both position-aware and position-free customization as in our approach. Moreover, unlike our framework, ACE++ does not incorporate the innovative ICMA module.  
For a fair comparison on \OurBench{}, we directly use ACE++’s publicly released \textbf{subject LoRA adapters} to evaluate its performance under our benchmark.  
As shown in Tab.~\ref{tab:comparisons-ace++} and Fig.~\ref{fig:comparisons-ace++}, our model consistently produces more identity-consistent and visually coherent customization results, showing superior perspective, lighting, and shape fidelity while operating as a single unified model rather than multiple task-specific LoRA adapters.

\begin{table}[t]
  \centering
  \small
  \captionsetup{skip=3pt}
  \caption{\textbf{Comparison with ACE++~\citep{mao2025ace++} on ProductBench.} Metrics under Precise Mask (left) and User-drawn Mask (right); higher is better ($\uparrow$).}
  \label{tab:comparisons-ace++}
  \setlength{\tabcolsep}{5pt}
  \renewcommand\arraystretch{1.1}

  \begin{minipage}[t]{0.48\linewidth}
    \centering
    \textbf{Precise Mask} \\
    \vspace{3pt}
    \begin{tabular}{lccc}
      \toprule
      \textbf{Method} & \textbf{DINO-I} & \textbf{CLIP-I} & \textbf{CLIP-T} \\
      \midrule
      ACE++ & 60.68 & 81.34 & 31.64 \\
      \rowcolor{blue!5}\textbf{Ours} & \textbf{63.14} & \textbf{81.92} & \textbf{31.75} \\
      \bottomrule
    \end{tabular}
  \end{minipage}\hfill
  \begin{minipage}[t]{0.48\linewidth}
    \centering
    \textbf{User-drawn Mask} \\
    \vspace{3pt}
    \begin{tabular}{lccc}
      \toprule
      \textbf{Method} & \textbf{DINO-I} & \textbf{CLIP-I} & \textbf{CLIP-T} \\
      \midrule
      ACE++ & 61.26 & 81.16 & 31.42 \\
      \rowcolor{blue!5}\textbf{Ours} & \textbf{63.28} & \textbf{81.95} & \textbf{31.80} \\
      \bottomrule
    \end{tabular}
  \end{minipage}
\end{table}

\section{Training Strategy: Mask Sampling and Augmentation}
\label{appendix:mask_sampling_augmentation}
To enhance model flexibility and robustness, we randomly sample mask types during training: 
position-aware masks with a probability of 0.6 and position-free masks with 0.4. 
Within the position-aware cases, we further draw user-drawn masks with 0.75 probability and precise masks with 0.25, assigning higher probabilities to harder tasks to provide more training iterations. 
In addition, we convert precise masks from Grounded SAM into user-drawn masks via standard image-morphology operations such as dilation, erosion, opening, and closing.
\section{Web Application}
\label{appendix:web_app}
\begin{figure*}[ht]
    \centering
    \includegraphics[width=1.\linewidth]{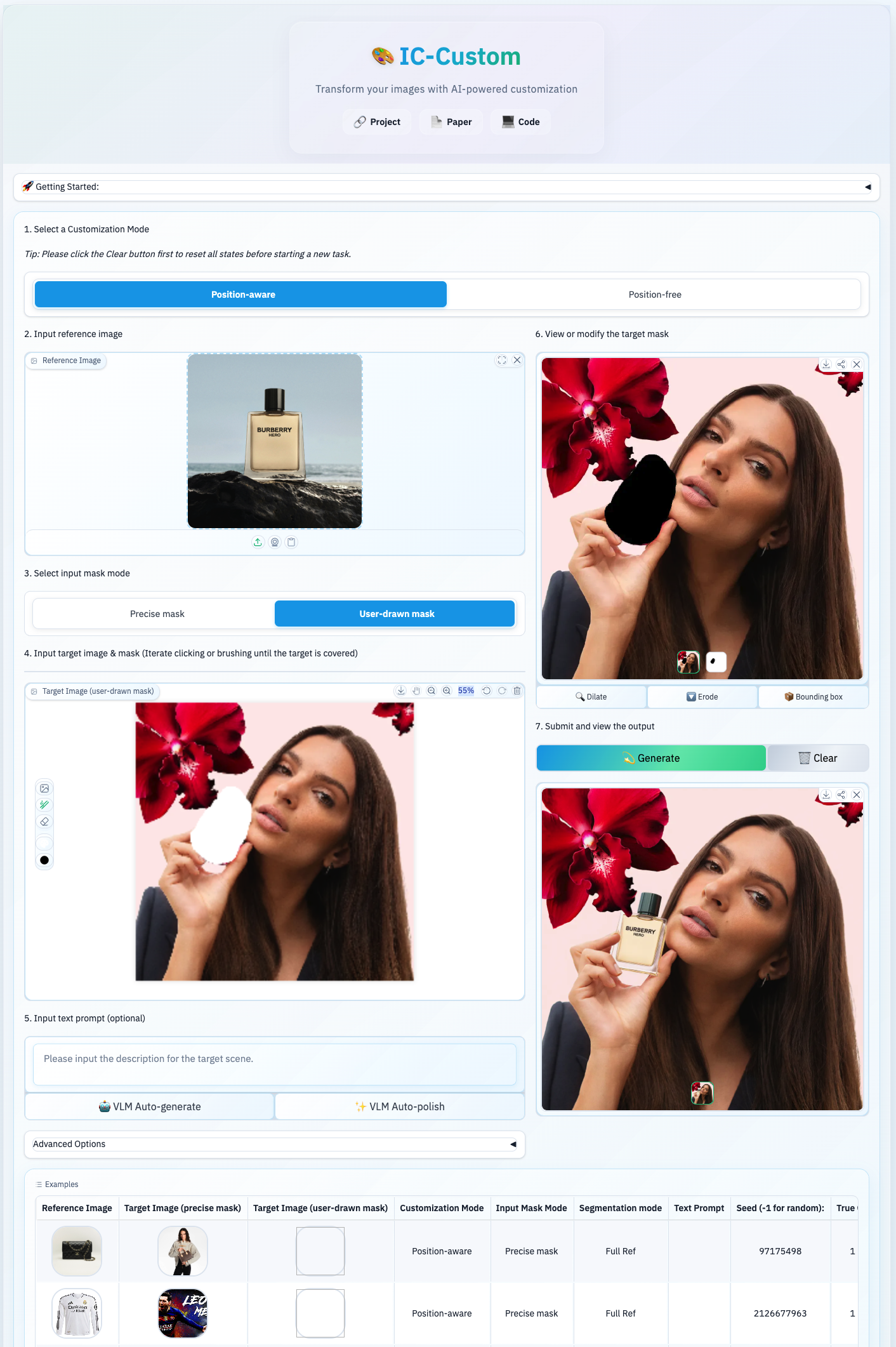}
\caption{\textbf{Web App – Position-aware mode.}  
Users upload a reference image and a fill-in image, choose the mask type (precise or user-drawn), optionally edit or refine the mask, add an optional text prompt, and then run the model to perform position-aware customization.}
    \label{fig:webAPP_pos_aware}
\end{figure*}

\begin{figure*}[ht]
    \centering
    \includegraphics[width=1.\linewidth]{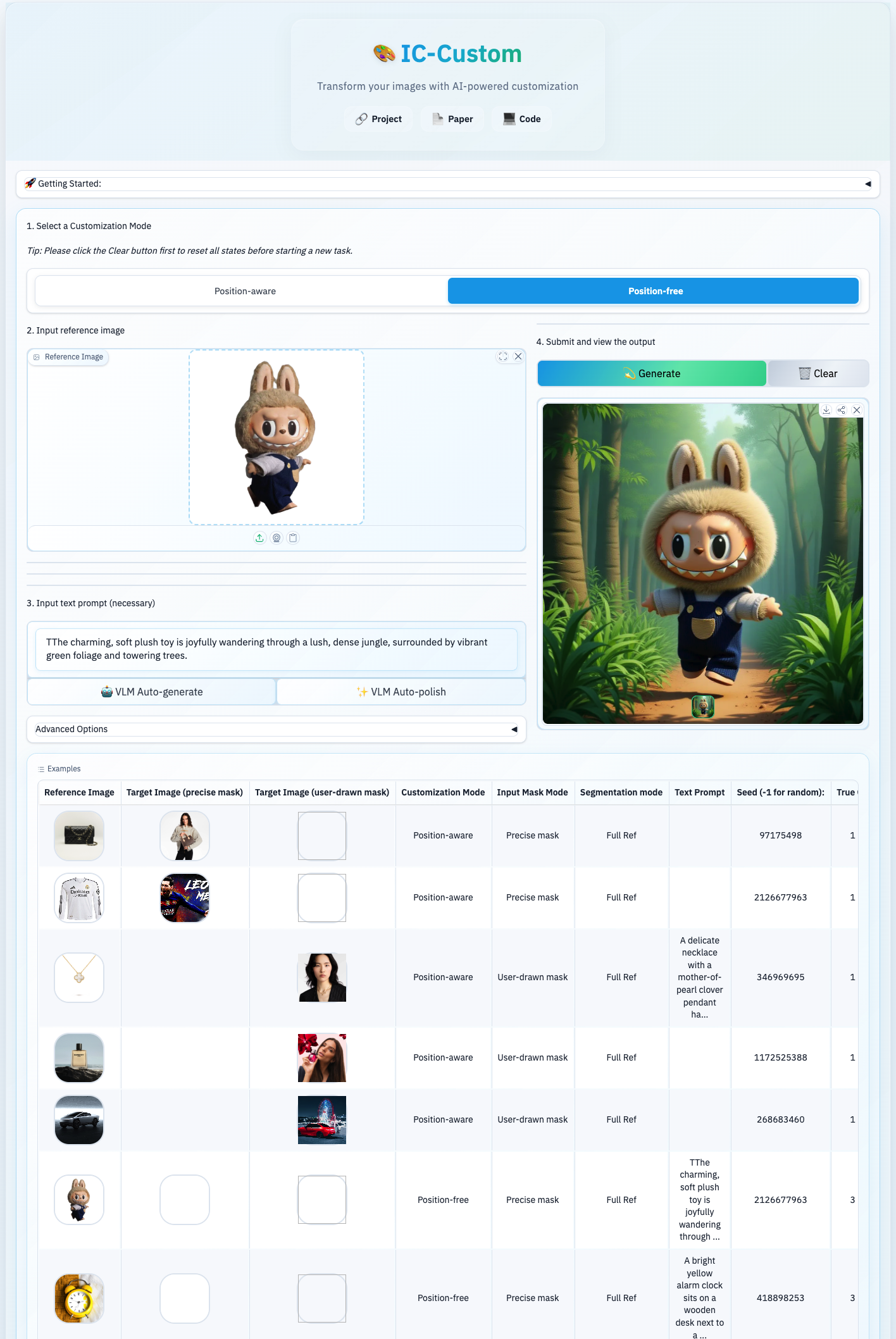}
\caption{\textbf{Web App – Position-free mode.}  
Users upload a reference image, provide a text prompt describing the desired scene or use the built-in VLM prompt generator, and then run the model to perform position-free customization.}
    \label{fig:webAPP_pos_free}
\end{figure*}

We implement a web application using Hugging Face Gradio~\footnote{\url{https://www.gradio.app/}} to provide a simple and seamless interface for both position-free and position-aware customization (see Fig.~\ref{fig:webAPP_pos_aware} and Fig.~\ref{fig:webAPP_pos_free}).  
Users first select a customization mode and upload a reference image.  
In the \textbf{position-aware mode}, they choose a mask type (precise or user-drawn), upload the fill-in image, optionally refine the mask (via SAM for precise masks or manual brushing for user-drawn masks), and provide an optional text prompt before running the model.  
In the \textbf{position-free mode}, users directly supply a text prompt describing the desired scene or use the built-in VLM-based prompt auto-generation tool prior to execution.
This web application provides a simple, unified interface for both position-aware and position-free customization, enabling users to interactively explore our model’s capabilities with minimal setup. We will release the full code and the web application as open source to support reproducibility and community adoption.

\begin{figure*}[htbp]
    \centering
    \includegraphics[width=1.\linewidth]{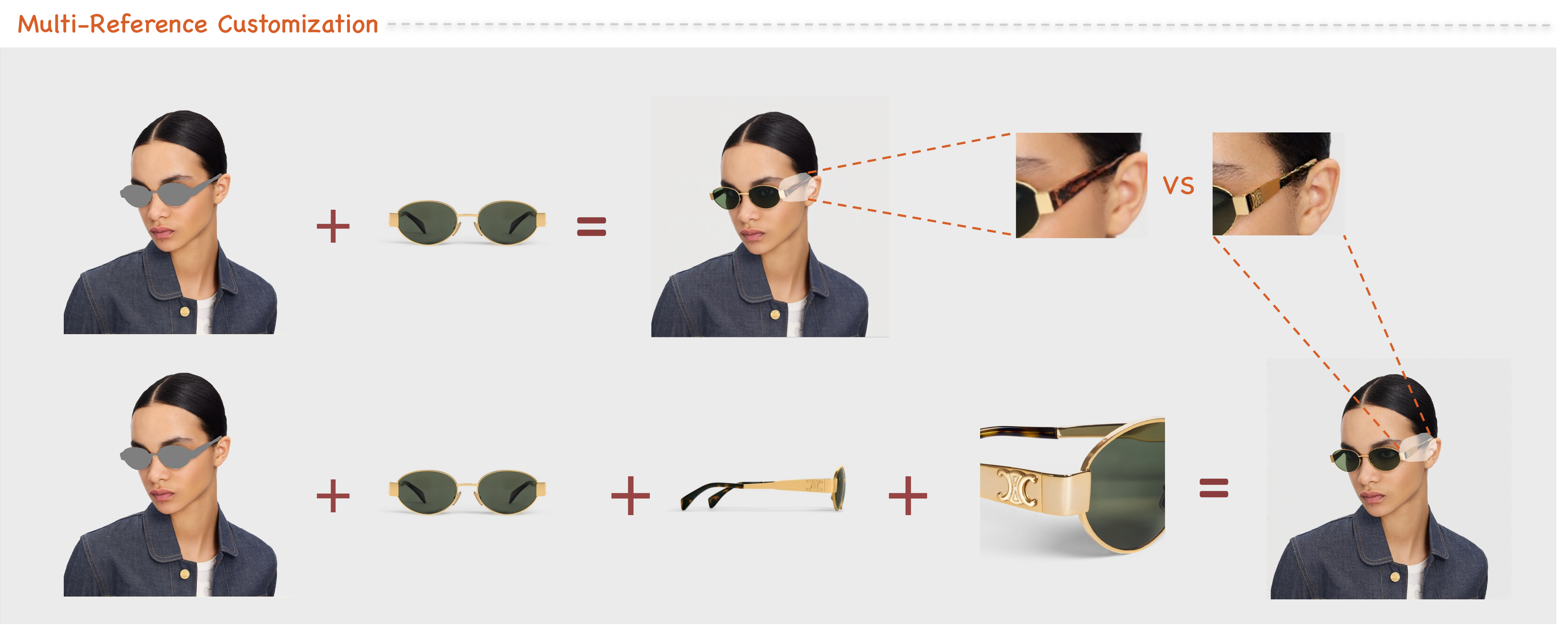}
\caption{\textbf{Multi-Reference Customization.}  
By aggregating multiple reference images of the same identity from different environments and viewpoints, our model preserves richer details and textures.  
For example, when a single reference view omits the glasses’ temples, the model must hallucinate them; with multiple viewpoints including the temples, it reconstructs the object more completely.}
    \label{fig:multi-ref}
\end{figure*}
\section{Preliminary Study on Multi-Reference Customization}
\label{appendix:multi-ref}
Benefiting from the learnable task-oriented register tokens and boundary-aware positional embeddings introduced in our \textbf{In-Context Multi-Modal Attention (ICMA)}, our model can accurately distinguish customization types and the boundaries between inputs and outputs. 
This naturally extends to \textbf{multi-reference customization}, where multiple reference images of the same identity (but from different scenes) are provided---not as multi-image fusion but as separate context cues.  
By aggregating information from multiple references, our model better preserves identity fidelity and fine details.  
To support this setting, we concatenate multiple reference images with the fill-in noise input and introduce an additional \textbf{index embedding} in the boundary-aware positional embeddings to differentiate reference indices.  
We also curated a multi-reference dataset containing \textbf{2K real-world} and \textbf{2K synthetic polyptychs} for training. 
As shown in Fig.~\ref{fig:multi-ref}, our multi-reference approach aggregates information from multiple references (e.g., different viewpoints) to better preserve object identity details and textures. This preliminary exploration highlights the broader capability and scalability of our unified customization model, and we plan to further explore this direction in future work.

\section{Additional Visualization Results}
\label{sec:more_visualization}
\begin{figure*}[htp]
    \centering
    \includegraphics[width=.92\linewidth]{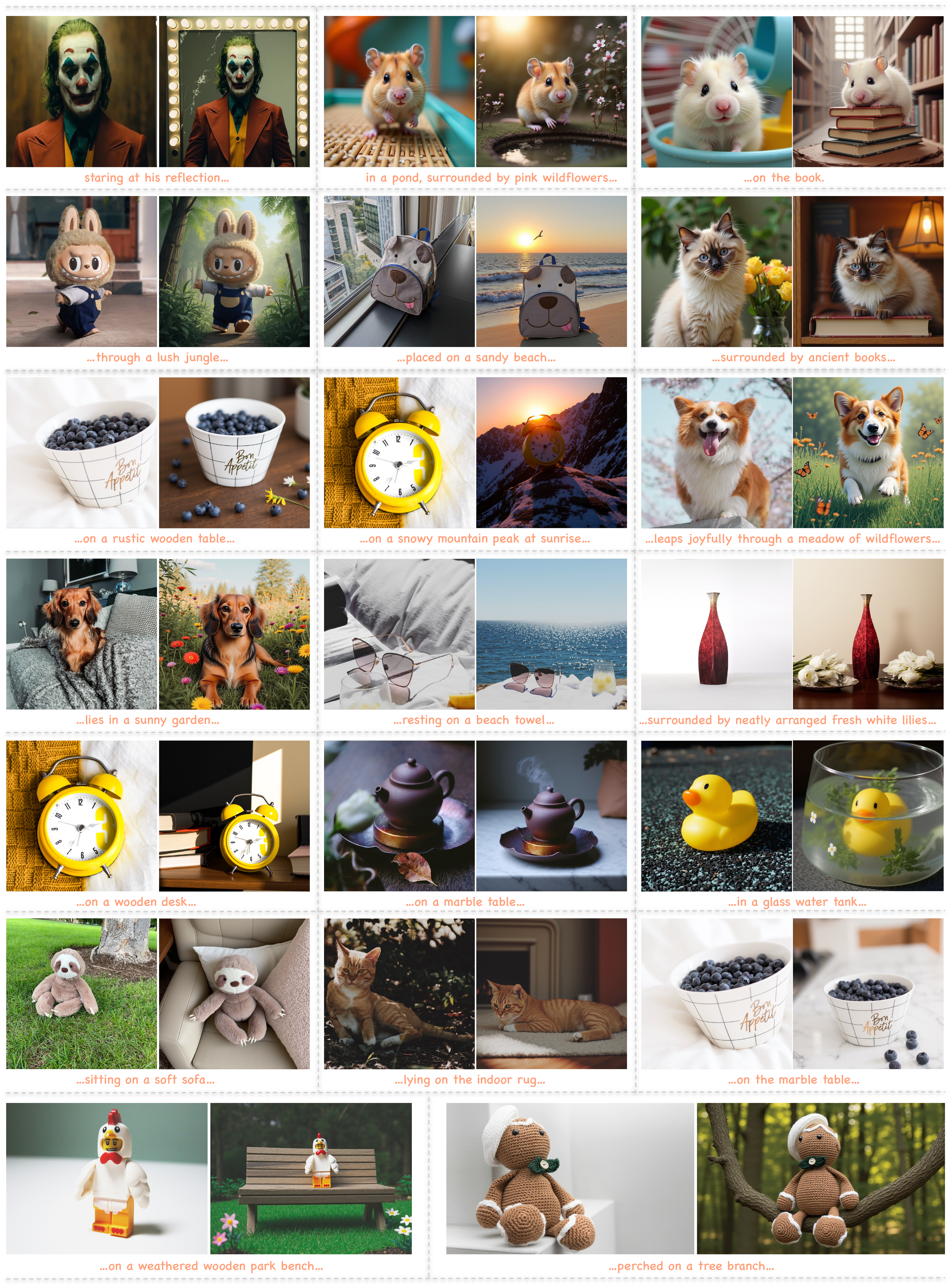}
    \caption{\textbf{Additional visualization results on position-free customization.} Our method successfully maintains identity consistency while generating diverse scenes and poses.}
    \label{fig:more-results-pos-free}
\end{figure*}

\begin{figure*}[htp]
    \centering
    \includegraphics[width=.92\linewidth]{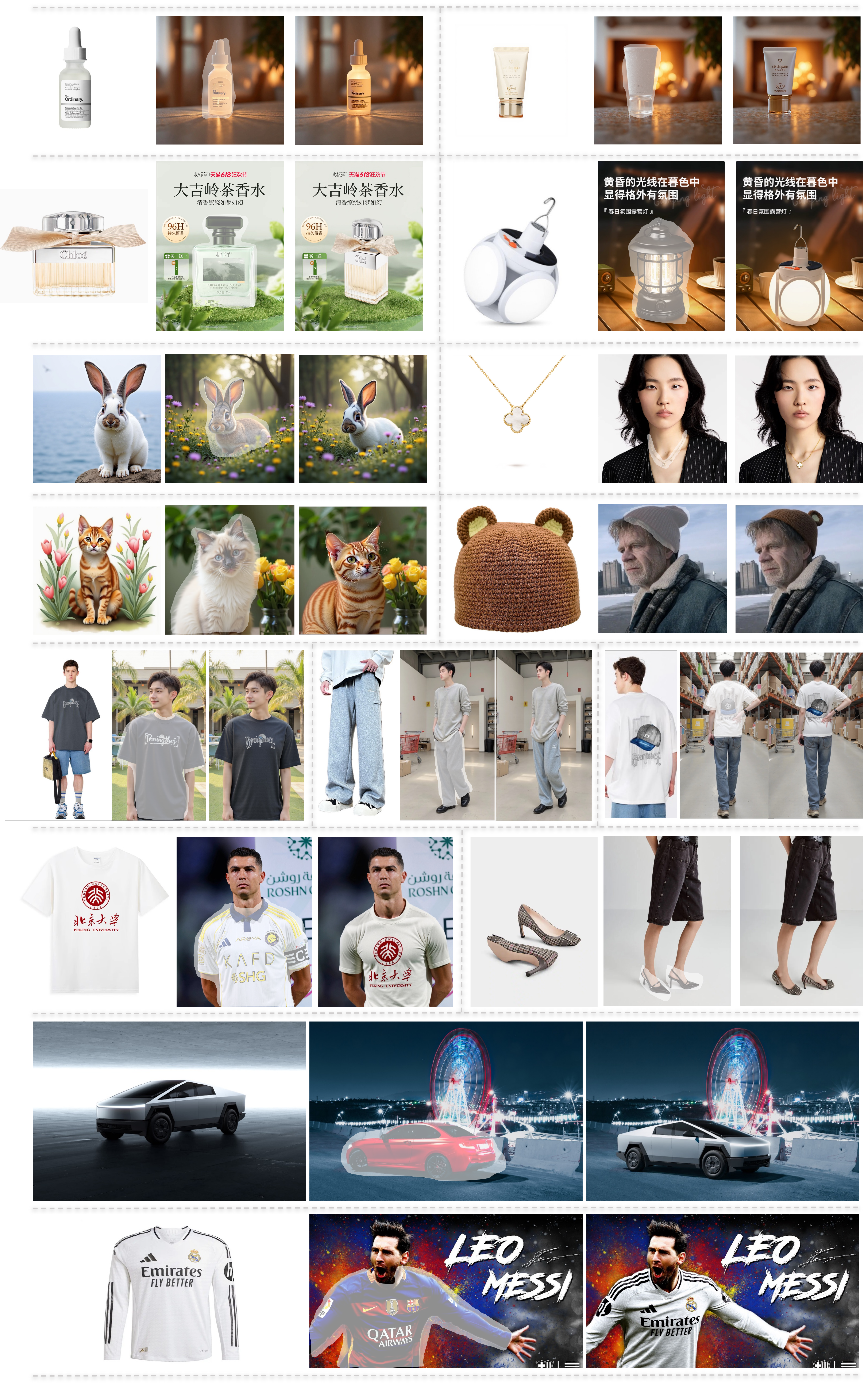}
    \caption{\textbf{Additional visualization results on position-aware customization.} Our method successfully maintains identity consistency while seamlessly integrating subjects into diverse lighting, styles, and poses in target scenes.}
    \label{fig:more-results-pos-aware}
\end{figure*}

Figure~\ref{fig:more-results-pos-free} shows additional position-free customization results, where our model seamlessly generates novel scenes that preserve the reference identity based on text descriptions.
Figure~\ref{fig:more-results-pos-aware} presents additional position-aware customization results, demonstrating its ability to accurately insert or edit images with different materials and textures while maintaining identity consistency.
\section{Ethics Statement}
This work complies with the ICLR Code of Ethics.\footnote{\url{https://iclr.cc/public/CodeOfEthics}}  
Our study does not involve human or animal subjects, personally identifiable information, or sensitive demographic attributes. All datasets are either publicly available or internally curated, and will be verified for proper licensing prior to open-sourcing. We also adopt the SafeChecker from the Diffusers FLUX.1 framework to filter potentially harmful outputs (e.g., sexual, violent, or toxic content) and apply similar precautions during data collection to minimize such content. We adhere to established research integrity practices, including reproducibility, transparency, and proper attribution of prior work.

\section{Reproducibility Statement}
To ensure reproducibility, we provide detailed descriptions of our data preparation and processing in Sec.~\ref{sec:in-context-customization-data-curation}, and implementation details in Sec.~\ref{sec:expriments_setup}, including training hyperparameters, evaluation protocols, and baselines clarification. In Appendix Sec.~\ref{sec:automated_caption}, we also describe the prompts used when preparing data with the multi-modal language model. We will release our code and models under appropriate licenses to facilitate full reproducibility.

\section{LLM Usage Statement}
In preparing this paper, we used large language models (LLMs), including ChatGPT~\citep{gpt4o} and Gemini~\citep{team2023gemini}, solely as writing-assistance tools. Specifically, we first drafted the content ourselves and then used LLMs with prompts such as “You are an expert in academic writing. Please help me refine and rephrase the text to make it more professional, fluent, clear, and readable.” We then manually reviewed and revised all LLM outputs to ensure that the text accurately reflects our intended meaning.  
No part of the research design, experiments, analysis, or results was generated by LLMs; their use was limited to improving clarity and readability of the manuscript. We, the authors, take full responsibility for the content of this paper.



\end{document}